\documentclass[journal]{IEEEtran}
\usepackage{microtype}
\usepackage{graphicx}
\usepackage{booktabs}
\usepackage{xcolor}

\usepackage{array}
\usepackage[caption=false,font=normalsize,labelfont=sf,textfont=sf]{subfig}
\usepackage{textcomp}
\usepackage{dblfloatfix}
\usepackage{url}
\usepackage{verbatim}
\usepackage[numbers]{natbib}
\usepackage{amsmath}
\usepackage{amssymb}
\usepackage{mathtools}
\usepackage{amsthm}

\theoremstyle{plain}
\newtheorem{theorem}{Theorem}[section]
\newtheorem{proposition}[theorem]{Proposition}
\newtheorem{lemma}[theorem]{Lemma}
\newtheorem{corollary}[theorem]{Corollary}
\newtheorem{observation}[theorem]{Observation}
\theoremstyle{definition}

\theoremstyle{remark}

\newtheorem{setting}{Setting}

\newcommand{\A}{\mathsf{A}}
\newcommand{\G}{\mathsf{G}}
\newcommand{\M}{\mathsf{M}}
\newcommand{\E}{\mathbb{E}}
\newcommand{\op}{_{\mathrm{op}}}

\usepackage{algorithm}
\usepackage{algorithmic}
\newif\ifLong
\Longfalse

\providecommand{\IEEEPARstart}[2]{#1#2}

\newcommand{\appref}[1]{\ifLong Section~\ref{#1}\else Appendix~\ref{#1}\fi}

\begin{document}

\Longtrue
\pagestyle{plain}

\title{Spectrum Is Not Enough: A Random-Matrix Account of\\
Data-Free Preconditioning for Differentially Private Learning}

\author{Shilin Zhang and Kangning Wang\\[2pt]
\small School of Electrical and Control Engineering, North China
University of Technology, Beijing, China\\[1pt]
\small Corresponding author: Shilin Zhang, \texttt{zhangshilin@126.com}, ORCID 0000-0003-3034-1538}

\markboth{Spectrum Is Not Enough}%
{Zhang and Wang}

\maketitle

\begin{abstract}
At a fixed privacy budget, curvature is exactly the object the budget makes
unavailable. A recent answer estimates it from synthetic probes, at neither
privacy cost nor distribution shift, and validates a probe by the eigenvalue
decay it reproduces. That test reads the spectrum, not the basis, and what a
probe reconstructs is unknown. This paper gives a random-matrix account of
data-free preconditioning, organized by one dichotomy. Of the two
Kronecker factors, the error-signal factor $\G$ and the
architecture-determined bulk of the activation factor $\A$ are reconstructible
without data up to an explicit error-energy bound; only the data-induced spikes
of $\A$ are not. We verify that bound with zero violations along a full
training trajectory, though it stays one to nearly three orders of magnitude
above the gap it bounds. We locate the failure boundary at a
Baik--Ben Arous--P\'ech\'e threshold: a bulk-only probe suffers an exact damage
$s^2/(1-s^2)$ from a spike of strength $s^2$, invisible once its amplitude
exceeds the scaled-bulk threshold $(\sqrt{c}/(1+\sqrt{c}))^{1/2}$, $c = d/n$.
\par\smallskip\noindent
The test in use is basis-blind: rotating only the probe eigenbasis
at a near-exact matched spectrum inflates the residual by $2.3\times$ to
$89.8\times$, and a closed-form volume law fixes the within-spectrum ordering of
the critical volume while a two-oracle construction proves that no functional
of the probe spectrum fixes its magnitude. Over forty seeds on three
modalities, per-layer tuning does
not transfer to accuracy, no data-free arm beats the flat probe, public-data
curvature does on both modalities tested, and the ordering inverts on
PathMNIST under real DP noise. The account yields a
probe-design protocol and falsifiable predictions for the regimes where
data-free curvature can be trusted.
\end{abstract}


\section{Introduction}
\label{sec:intro}

\IEEEPARstart{D}{ifferential} privacy \citep{dwork2014} is the standard formal
guarantee for training on sensitive data, and DP-SGD \citep{abadi2016} its
workhorse. Adoption is limited by a geometric conflict: neural losses are
strongly ill-conditioned \citep{dauphin2014}, while DP-SGD clips per-sample
gradients and injects spherical noise, treating all parameter directions
equally. Preconditioning is the standard remedy, and at a fixed privacy budget
the utility a run reaches depends on how well the optimizer matches the loss
geometry, so a curvature estimate that costs no privacy budget buys utility at
fixed $\varepsilon$; the same access limit governs second-order estimation in
federated and continual settings. Second-moment statistics from private data
are sensitive, though: privatizing a full preconditioner is noisy and often
unstable, adaptive schemes that chase unbiased curvature can underperform
their non-adaptive counterparts \citep{ganesh2025}, and federated second-order
information adds communication bottlenecks \citep{krouka2025}. A third line
approximates the geometry on public data \citep{amid2022}, at the cost of
distribution shift and audit complications.

DP-KFC \citep{dpkfc2026} proposes a data-free route: estimate the KFAC factors
\citep{kfac2015} from synthetic probes that carry no privacy cost at all,
using colored noise shaped to a $1/f^{\alpha}$ spectrum, motivated by the
scale-free statistics of natural images. The recipe validates its probes
empirically: the synthetic factors must recover the eigenvalue decay of the
private ones. That test reads only the spectrum of the factors and ignores the
eigenbasis, and it treats the factorization as one object, although the two
Kronecker factors are built from different signals: the error signals the
architecture propagates, and the activations of the data. It is therefore
unknown whether any part of the factorization is data-free by construction,
and where a spectrum-matched probe begins to fail.

In this paper we answer with the tools of random matrix theory, the machinery
that governs sample covariance spectra at large dimension
\citep{marchenko1967,johnstone2001}: deformed bulks and spiked covariance
models whose data-induced outliers detach at the Baik--Ben Arous--P\'ech\'e
threshold \citep{baik2005}, building on signal propagation through spiked
covariances in deep networks \citep{wang2024}. The private activation
covariance is a deformed bulk plus a few data-induced spikes, and the
error-signal factor is a weighted sum of the activations the network
propagates, so the two factors differ in one respect: which of their
components the data drives. We organize the account by a single statement, a reconstructibility
dichotomy: the error-signal factor $\G$ is source-insensitive, and inside the
activation factor $\A$ the architecture-determined bulk is recoverable while
the detached data-induced spikes are not. What a probe must carry for the bulk
is the basis, not the spectrum. Four consequences follow.

\textbf{A proof that the error-signal half is data-free by nature}
(Section~\ref{sec:pillarG}). KFAC approximates the curvature as a Kronecker
product $\A \otimes \G$ of an activation factor $\A$ and an error-signal
factor $\G$. We prove an operator-norm bound stating that $\G$ varies between
any two data sources by at most an error-energy term that vanishes as the
model's mean true-class confidence approaches one. The bound is calibrated
across three source types, four layers, and six training snapshots
($72$ source--layer--epoch configurations), with zero violations (Section~\ref{sec:calib}).

\textbf{A falsifiable failure boundary} (Section~\ref{sec:pillarT2}). We model
the private activation covariance as a spiked covariance: a deformed bulk
determined by the architecture, plus data-induced spikes. A bulk-only probe
suffers an exact, monotone operator-norm damage of $s^2/(1-s^2)$ from a spike
of strength $s^2$, and the BBP threshold identifies when the spike detaches
from the bulk and becomes invisible to any data-free probe. The exact
scaled-bulk threshold is $s_c = (\sqrt{c}/(1+\sqrt{c}))^{1/2}$, which at
$d/n = 0.191$ gives $0.551$ against $0.661$ for the leading-order $c^{1/4}$
and matches a crossing-based estimate on an injected-spike surrogate.

\textbf{A mechanistic correction, with the spectrum ruled out as the operative
object} (Sections~\ref{sec:anatomy}--\ref{sec:downstream}). Through experiments
that hold the probe spectrum fixed and vary only its eigenbasis, we show that
spectral shape matching is neither sufficient for, nor the operative object
of, preconditioner quality: eigenvalue diagnostics become actively misleading
in steep regimes, and the trace of the residual under basis misalignment is
set by the whitened volume of the misaligned subspace, with the measured
anisotropy following the same ordering but not the same magnitude. We derive
that volume in closed form, give its partial-volume version for a rotated prefix, and prove that the
prefix volume grows by at least one unit per added direction, which orders the
critical volume by spectral steepness and identifies the white probe as the
unique basis-immune spectrum. A two-oracle construction then shows that the
magnitude of the damage is not a function of the spectra at all, so the title
claim is a theorem and not a summary of the experiments.

\textbf{The verdict on the data-free premise under a controlled design}
(Sections~\ref{sec:downstream} and~\ref{sec:negative}). Forty seeds per
arm on three modalities, with the private oracle as an upper reference: no
data-free arm separates from the flat probe, against a headroom of $+1.9$ to
$+6.8$ points that the oracle reaches, so per-layer operator tuning buys no
accuracy and the residual orders sources without ordering magnitudes.
Public-data curvature, which costs no privacy budget either, does separate on
both modalities tested. Under a real privacy mechanism the source ordering
survives while learning survives and then inverts, the flat probe overtaking
the matched ones once the noise removes the structure they matched, so the
bottleneck moves from preconditioning to learnability, and two measurement
pitfalls bound what the theory can claim. Figure~\ref{fig:concept}, in
\appref{app:overview}, summarizes the setting, the inversion, and the volume law.

\ifLong
\textbf{Organization.} Section~\ref{sec:setting} fixes the notation and the
experimental platform, and Section~\ref{sec:related} places the work in the
literature. Sections~\ref{sec:pillarG} and~\ref{sec:pillarT2} prove the two
pillars: the error-signal invariance and the exact detachment boundary.
Section~\ref{sec:anatomy} isolates the basis-first mechanism and its closed
form, Sections~\ref{sec:downstream} and~\ref{sec:negative} follow the account
downstream and under a real privacy mechanism, Section~\ref{sec:protocol}
condenses it into a protocol, and Section~\ref{sec:discussion} states what
survives the privacy wall. Proofs and the full experiment records are in
Sections~\ref{app:proofs} and~\ref{app:records}.
\fi

\section{Setting}
\label{sec:setting}

\begin{setting}[KFAC factors under the data-free recipe]
\label{set:kfac}
Fix a network $f_\theta$ and a loss $\ell$. For a linear layer
$W_l \in \mathbb{R}^{d_l \times d_{l-1}}$ (bias appended as an extra
ones-column in the activation), let $a_l(x) \in \mathbb{R}^{d_{l-1}+1}$ denote
the input activation and $\delta_l(x) \in \mathbb{R}^{d_l}$ the output
gradient $\partial \ell / \partial s_l(x)$. Given a source $S$ over
input--label pairs, the KFAC factors are
\begin{equation}
\begin{aligned}
\A_l(S) &= \E_{(x,y)\sim S}\!\left[a_l(x)a_l(x)^{\top}\right] + \varepsilon I,\\
\G_l(S) &= \E_{(x,y)\sim S}\!\left[\delta_l(x)\delta_l(x)^{\top}\right] + \varepsilon I ,
\end{aligned}
\end{equation}
estimated over $n$ samples of $S$. The data-free recipe replaces $S$ by a
probe ensemble $P$ and preconditions the gradient of $W_l$ as
$\G_l(P)^{-1/2}\, \nabla_{W_l}\ell\; \A_l(P)^{-1/2}$, applied before the
privacy mechanism.
\end{setting}

\textbf{Random-matrix background.} For an isotropic population and aspect
ratio $d/n$ the sample covariance spectrum concentrates on the
Marcenko--Pastur law \citep{marchenko1967}, and a rank-one population spike of
strength $\beta$ detaches from the edge if and only if $\beta > \sqrt{d/n}$
\citep{baik2005,johnstone2001}. ``Bulk'' denotes the spectral ensemble
excluding finitely many data-induced outliers and ``spike'' those outliers,
the decomposition of trained-network spectra in \citet{sagun2016}, formalized
in Lemma~\ref{lem:bbp}. The metric is the residual
$\M = \A_p^{-1/2}\A_o\A_p^{-1/2}$ of oracle $o$ against probe $p$. With
$\lambda_1 \le \dots \le \lambda_d$ its eigenvalues and $\bar\lambda$ their mean,
the total anisotropy is $\mathcal{A} = \sum_i(\lambda_i-\bar\lambda)^2 /
(d\bar\lambda^2)$, the spike share is $s_1 = (\lambda_d-\bar\lambda)^2 /
\sum_i(\lambda_i-\bar\lambda)^2$, and the bulk residual is
$bulk_1 = \mathcal{A}(1-s_1)$, the anisotropy left once the leading direction is
removed, so a probe that leaves only sampling noise sits near $1$. The whitening
clamps probe eigenvalues at $10^{-2}$, and every comparison below holds the
trained model fixed. A log-ratio dispersion over the leading eigenvalues of the
paired spectra measures spectral shape matching, and a tail exponent $\beta$ captures
steepness.

\textbf{Experimental platform.} We use eight dataset configurations: four
MedMNIST modalities \citep{medmnist2023,medmnist2021} (PathMNIST, OCTMNIST,
DermaMNIST, TissueMNIST), a grayscale $32\times32$ CIFAR-10 subset, the full
$50$k CIFAR-10 training set, CIFAR-100 restricted to ten classes
\citep{krizhevsky2009}, and DermaMNIST at $64\times64$ RGB. The two
architectures are a tanh MLP and a two-convolution-layer network whose KFAC
factors use an unfold approximation (\appref{app:records}). Every classifier
uses the same ten-way output head, so the last-layer error factor is comparable
across datasets and balanced accuracy is macro recall over those ten slots, in
which a constant predictor scores $1/10$ and uniformly random labels $|C|/100$. Training is
per-sample-clipped DP-SGD with per-epoch factor re-estimation. Twenty
experiment families support the claims; every number below is tied
to a recorded run, and each printed cell is recomputed from the recorded
result files by a script.

\ifLong
\section{Related Work}
\label{sec:related}

\textbf{Preconditioning under differential privacy.} Curvature under DP comes
from three routes: public-data proxies \citep{amid2022}, private curvature
estimators whose second moments must themselves be privatized
\citep{ganesh2025,krouka2025}, and data-free synthetic probing
\citep{dpkfc2026}, which is the setting analyzed here. DP-KFC validates a probe
by the eigenvalue decay it recovers; we turn that criterion into a
reconstructibility statement with a closed-form failure boundary.

\textbf{Random-matrix accounts of network spectra.} The bulk-plus-spike
picture of trained-network covariance \citep{sagun2016} has been extended to
exact signal propagation of nonlinear spiked models \citep{wang2024} and,
broadly, to the deterministic- and high-dimensional-equivalent frameworks
\citep{couillet2022,liao2026} that summarize the proportional-regime analysis
of linear and nonlinear models. Ours is the same machinery one level down, at
a working preconditioner rather than a kernel: the private activation factor
deforms an architecture-determined bulk, the BBP transition
\citep{baik2005,johnstone2001} separates a reconstructible bulk from an
irrecoverable spike, and detection-theoretic formulations of spiked models
\citep{jung2023} address the identifiability side of the same question. The
Kronecker structure of the Fisher block is itself the object of recent
random-matrix work on Kronecker-product random matrices \citep{fan2026},
which gives free-deterministic-equivalent approximations for exactly the
$\A \otimes \G$ combination the data-free recipe forms.

\textbf{Whiteners and probability-level inference.} Beyond Kronecker-factored
curvature, matrix-aware optimizers whiten the two gradient factors (the
Shampoo and Muon families); recent work analyzes when coordinate whitening
must precede spectral whitening \citep{zeta2026}, and mean-activation
approximations of the two KFAC factors are analyzed in \citep{mac2025}. Our
results transfer to these only in the form of a prediction
(Section~\ref{sec:discussion}), because the basis-first law so far has been
verified on Kronecker-factored preconditioners. At the privacy end, the
asymptotic variance of DP-SGD decomposes into statistical, sampling and
privacy-induced components \citep{xia2025}; the noise cross-over of
Section~\ref{sec:negative} is the preconditioning-level counterpart.
\fi

\section{Pillar I: the error-signal factor is data-insensitive}
\label{sec:pillarG}

\begin{proposition}[Error-energy sensitivity bound]
\label{prop:g}
Let $S, S'$ be two sources and fix a model $f_\theta$ and layer $l$. Then:
\begin{enumerate}
\item[(i)] $\big\|\E_S[\delta\delta^{\top}] - \E_{S'}[\delta'\delta'^{\top}]
\big\|\op \;\le\; \E_S\|\delta\|^2 + \E_{S'}\|\delta'\|^2$.
\item[(ii)] At the last layer with cross-entropy loss,
$\delta = p(x) - e_{y}$ where $p(x)$ is the softmax vector and $e_y$ the
one-hot label, hence
$\E\|\delta\|^2 \le 2(1 - \E\, p_y)$. If the mean true-class confidence of
$f_\theta$ is at least $1-\eta$ on both sources, then
$\|\G_l(S) - \G_l(S')\|\op \le 4\eta$: the two $\varepsilon I$ terms are the
same constant and cancel in the difference, so the bound is independent of the
modalities of $S$ and $S'$.
\item[(iii)] No modality term enters, but the bound is not tight: on the
calibration grid the measured gap is at most $6.4\%$ of the right-hand side and
$0.15\%$ at the median. At initialization, where $p(x)$ is uniform for
standardized inputs, $\eta \approx 1-1/C$ and it is vacuous outright.
\end{enumerate}
\end{proposition}

The proof is elementary and appears in \appref{app:proofs}. Part (i) is loose
by one to nearly three orders of magnitude: the measured sensitivity never
exceeds $6.4\%$ of it
(Table~\ref{tab:calib}), so it rules out large source effects but does not
pin them. Part (ii) is stated for the last layer only, where
$\delta = p - e_y$ holds exactly; its extension to hidden layers is empirical.
The operative content is therefore not tightness but the measured fact below:
the $\G$ residual anisotropy is constant across sources. Proposition~\ref{prop:g}
licenses the data-free recipe on the $\G$ half of the Kronecker product for
any probe, and the calibration below finds that constancy at every stage,
including initialization, where the bound itself is vacuous, so it is expected
rather than accidental. Empirically the \emph{shape} of the $\G$ spectrum
(indeed its residual anisotropy after preconditioning) is constant across
sources while its \emph{scale} drifts: the form admits overall rescaling but
forbids shape distortion. This reading is used quantitatively in
Section~\ref{sec:discussion}.

\subsection{Calibration}
\label{sec:calib}

The bound holds everywhere along the training trajectory (snapshots at epochs
$0$ to $16$, confidence parameter $\eta$
moving from $0.904$ down to $0.655$, $72$ source--layer--epoch
configurations): both bounds of the proposition are satisfied with zero
violations, and the actual sensitivity never exceeds $6.4\%$ of the (i) bound,
with a median of $0.15\%$ (Table~\ref{tab:calib}). What
changes across training is not the shape of $\G$ but its overall scale, which
tracks the loss; the residual anisotropy itself drifts only with the training
stage ($0.0057 \to 0.0067$ on PathMNIST). Across a sweep of four
biomedical modalities $\times$ two architectures $\times$ three source types
the largest per-layer $\G$ anisotropy is $0.096$. The mean-field variance
contraction that sets the bulk scale of $\A$ is likewise source-independent up
to $O(1/w)$ corrections, an immediate consequence of standard signal
propagation theory \citep{poole2016,schoenholz2017}; the measured fixed-point
discrepancy between sources is $2.6\%$. Since one of the two factors is
insensitive to the probe choice by construction, all accuracy-relevant mismatch
must live in $\A$.

\section{Pillar II: the failure boundary is a BBP threshold}
\label{sec:pillarT2}

\begin{lemma}[Exact detachment threshold under a scaled bulk]
\label{lem:bbp}
Let $a = \sqrt{1-s^2}\, z + s\, g\, u$ with $z \sim \mathcal{N}(0, I_d)$,
$g \sim \mathcal{N}(0,1)$, fixed unit $u$, and
$\A = \frac1n\sum_i a_i a_i^{\top}$. The population covariance is
$(1-s^2)I_d + s^2 u u^{\top}$: a rank-one spike of strength $\beta = s^2$ over
a bulk of variance $1-s^2$. With $c = d/n$, the sample top eigenvalue detaches
from the Marcenko--Pastur edge if and only if
\begin{equation}
\frac{s^2}{1-s^2} > \sqrt{d/n},
\qquad
s_c = \Bigl(\tfrac{\sqrt{c}}{1+\sqrt{c}}\Bigr)^{1/2}.
\end{equation}
The leading-order form $s_c = c^{1/4}$ over-states the threshold; at
$c = 0.191$ the exact value is $0.551$ against $0.661$ for the leading form
(Figure~\ref{fig:boundary}a).
\end{lemma}

The proof is one normalization step and appears in
\appref{app:proofs}. This is the Baik--Ben Arous--P\'ech\'e transition
for rank-one spiked sample covariance models \citep{baik2005}, in the regime
formalized for principal component analysis by \citet{johnstone2001}. The
reading for the data-free recipe is structural. The bulk of $\A$ is
architecture-determined: it is reconstructible from probes because it is not
about the data. A detached spike is data-induced: it is, by construction,
invisible to any probe that carries no information about the private data, and
the loss it causes cannot be recovered by better probe design.

\begin{proposition}[Spike damage and its visibility threshold]
\label{prop:t2}
Consider the spiked population covariance of Lemma~\ref{lem:bbp} and any probe
whose factor is blind to the spike, $\A_p = (1-s^2)I_d$. Then:
\begin{enumerate}
\item[(i)] the residual
$\M = \A_p^{-1/2}\A_o\A_p^{-1/2} = I_d + \frac{s^2}{1-s^2}\, uu^{\top}$
has top eigenvalue $1/(1-s^2)$: the damage a bulk-only probe suffers from an
unseen spike is monotone in $s$, basis-free, and exact.
\item[(ii)] whether the data-induced spike is visible in the sample at all is
the BBP question of Lemma~\ref{lem:bbp}: the sample top eigenvalue detaches
from the Marcenko--Pastur edge if and only if $s^2/(1-s^2) > \sqrt{d/n}$.
\end{enumerate}
The proof is in \appref{app:proofs}.
\end{proposition}

\begin{proposition}[General-bulk boundary]
\label{prop:t2bulk}
Let $\Sigma = \Sigma_b + \beta u u^\top$ with $\Sigma_b$ a bulk of limiting
spectral distribution $\mu_b$. The sample top eigenvalue has an outlier above
the bulk edge if and only if $\beta$ exceeds the bifurcation threshold of the
free-additive fixed point \citep{bg2011}; for a scaled-isotropic bulk
$\Sigma_b = \sigma_b^2 I$ this is $\beta_c = \sigma_b^2 \sqrt{c}$, recovering
Lemma~\ref{lem:bbp}. Numerically $\beta_c$ scales linearly with the population
bulk edge $b_+$ within a shape family ($\beta_c/b_+ \approx 0.52$ isotropic,
$\approx 2.21$--$2.24$ for power-law bulks of exponent $1$ and $2$;
Figure~\ref{fig:boundary}c), so the boundary is a functional of the bulk
spectrum and not of its edge alone. Under the sample edge
$\sigma_b^2(1+\sqrt{c})^2$ the isotropic ratio is instead
$\sqrt{c}/(1+\sqrt{c})^2 = 0.22$. The proof reduces to \citet{bg2011}
(\appref{app:proofs}).
\end{proposition}

\ifLong
\begin{figure*}[!t]
\centering
\includegraphics[width=0.98\textwidth]{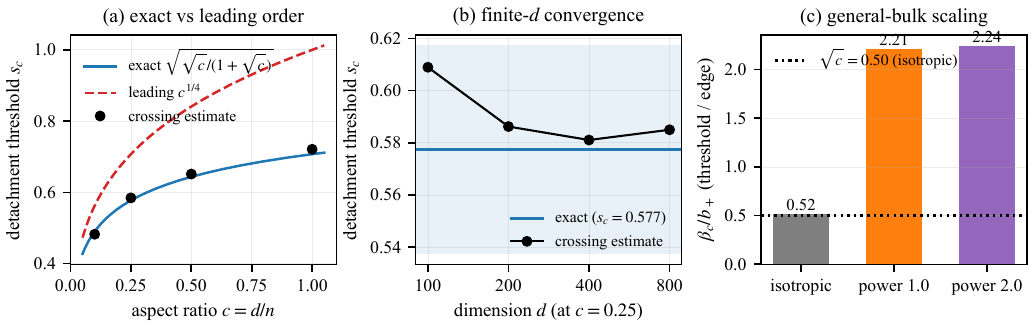}
\caption{Refinement of the failure boundary. (a) The exact threshold
$s_c = \big(\sqrt{c}/(1+\sqrt{c})\big)^{1/2}$ tracks the crossing-based
empirical estimate across aspect ratios, while the leading-order form
$c^{1/4}$ overstates it. (b) The empirical threshold converges to the exact
value as the dimension grows at fixed $c = 0.25$. (c) For an anisotropic bulk
the threshold is a functional of the bulk spectrum: $\beta_c/b_+ \approx
0.52$ for an isotropic bulk (with $\sqrt{c} = 0.5$, $b_+$ the population edge)
and $\approx 2.21$--$2.24$ for power-law bulks, independent of the exponent. Numbers from
verify\_boundary\_law (MC, archived in
figs/boundary\_refine\_data.json).}
\label{fig:boundary}
\end{figure*}
\fi

Experimentally, at $d/n = 0.191$ the asymptotic threshold of
Lemma~\ref{lem:bbp} is $s_c = 0.551$; under a finite detection margin the
transition is only flagged at $s \approx 0.7$, and a crossing-based estimate
recovers $0.551$ to within a few percent (Figure~\ref{fig:boundary}a,b). Below
the asymptotic threshold the top-1 share of the residual stays at the bulk
level ($1.00$--$1.02$), at $s = 0.75$ it reaches $1.26$, and by $s = 1$ the
residual is rank-one (Table~\ref{tab:bbp}). Trained activation spectra are not
Wishart-like, and that is a scope condition on the ruler rather than a
peculiarity of our probe: across the fifteen real factors of
\appref{app:records} the second eigenvalue sits $5.3$ to $73$ times above the
edge implied by the bulk mean, where on the injected model the same statistic
reproduces the Marcenko--Pastur edge to $1\%$. A threshold that presumes the
bulk shape therefore has no calibrated null on real factors, and a spike
strength relative to $s_c$ is not defined for them; the transition is mapped on
a surrogate because it is the only ensemble whose bulk the threshold describes. The practical
consequence is a division of labor: probes own the bulk; the DP clipping
mechanism owns the spike, because clipping bounds the influence of any single
direction, which is exactly the spike's failure mode.

\section{Anatomy of the bulk residual: basis first, spectrum second}
\label{sec:anatomy}

Propositions~\ref{prop:g} and~\ref{prop:t2} split the problem cleanly: all
probe-design freedom lives in the $\A$ bulk. DP-KFC's design heuristic is a
$1/f^{\alpha}$ probe family, motivated by the prevalence of scale-free
statistics in natural inputs. We tested this family systematically across
five modalities and depths $\{1,2,4\}$ and found that the family works, but
not for the advertised reason.

\subsection{Spectral shape matching is not the operative object}

\begin{observation}[Basis-first law]
\label{prop:basisfirst}
Separate the probe's eigenvalue spectrum from its eigenbasis and vary them
independently, at the input layer of trained models over five modalities:
\begin{enumerate}
\item[(i)] \textbf{Same spectrum, different basis.} A probe carrying the
exact private spectrum in the aligned Fourier basis stays at or below the
white-probe residual ($0.75$--$1.00$ of it across modalities); the same spectrum in a
Haar-random basis worsens the bulk-only residual by $2.3\times$
(PathMNIST) to $89.8\times$ (DermaMNIST), with the same eigenvalues and a
log-ratio dispersion that agrees to $10^{-6}$. A third arm carrying the oracle
eigenvalues in a Haar basis drives that dispersion down to $0.023$ and still
inflates the residual, against the aligned arm, by $2.2$--$84\times$.
Eigenvalue matching is neither
sufficient nor the operative object of the preconditioner residual.
\item[(ii)] \textbf{Misalignment is a step, not a curve.} Continuous rotation
$\theta \mapsto e^{\theta S}$ of the probe basis leaves the spectrum
invariant while the Fourier concentration of the probe collapses from
$0.31$--$0.44$ to $\approx 8.5 \times 10^{-3}$ already at $\theta = 0.3$,
the random-basis value; the bulk residual jumps by one to two orders of
magnitude over the same step and plateaus (Figure~\ref{fig:rotation} in \appref{app:records}). At
large $d$ any small random rotation is already Haar mixing, so there is no
partial-alignment regime.
\item[(iii)] \textbf{Damage is set by the whitened volume of the misaligned
subspace.} Rotating the top-$64$ eigendirections of the probe (carrying
$57$--$96\%$ of the probe variance) leaves the bulk residual unchanged on
every modality; a dense volume scan $K \in \{8, \dots, 512\}$ locates a
critical volume $K^{\star}$ that moves monotonically with the steepness of
the data spectrum (Figure~\ref{fig:records}(b), Table~\ref{tab:volume}). Beyond
$K^{\star}$ the damage grows superlinearly on steep spectra, saturating
toward the Haar-basis asymptote of (i), while rotation of the bottom-$512$
block is mild on every modality ($\le 1.2\times$).
\item[(iv)] \textbf{Tolerance scales with spectral steepness.} At rotation
scale $\theta = 1$, the residual relative to the aligned probe is $0.975$ on
the flattest modality ($\beta_o = -1.08$) but degrades $3.5\times$
(CIFAR-10), $5.3\times$ (TissueMNIST), $14.7\times$ (DermaMNIST) and
$16.7\times$ (OCTMNIST); $\theta \le 0.3$ is harmless on all modalities.
\item[(v)] \textbf{Eigenvalue diagnostics invert in the steep regime.} Over
$43$ non-flat layer-configurations on the extended grid
$\alpha \in [0,6]$, the layer-wise Spearman correlation between spectral
dispersion and the bulk residual is positive on $31$ and negative on $12$;
eight of the twelve negatives sit at input layers whose private bulk is
either too flat for any probe to resolve ($\beta_o \approx -1.08$) or
already steeper than the probe family can reach ($-2.23$), and the steepest
case inverts perfectly (Figure~\ref{fig:dissoc}).
\end{enumerate}
\end{observation}

\begin{figure*}[!t]
\centering
\includegraphics[width=\textwidth]{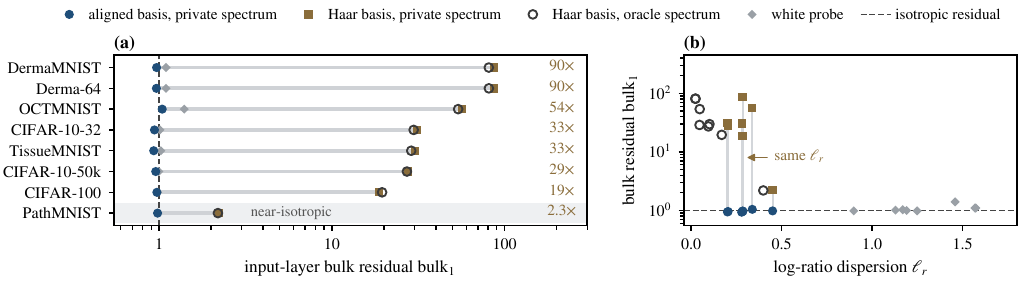}
\caption{The same eigenvalues in two bases. (a)~Input-layer bulk residual of
four probe arms over eight dataset configurations (mean of two probe bases per
arm). The probe that carries the private spectrum in the Fourier-aligned basis
stays at $0.75$--$1.00$ of the white-probe residual; the same spectrum in a
Haar basis inflates it by $2.3\times$ (near-isotropic PathMNIST, shaded) to
$89.8\times$. (b)~The same arms against the log-ratio dispersion, the
diagnostic the shape hypothesis recommends: the joined pair shares it to
$10^{-6}$ and differs in residual by up to $90\times$, and the arm carrying the
oracle spectrum in a Haar basis lowers it to $0.023$ while leaving the residual
at $74\times$ its white-probe value.}
\label{fig:basis}
\end{figure*}

Figure~\ref{fig:basis} places items~(i) and~(v) on one axis.

The next proposition gives the closed form behind items (i)--(iv).

\begin{corollary}[Eigenvalue profile is not a sufficient statistic]
\label{cor:sufficient}
The spectral invariants of the residual
$\M = \A_p^{-1/2}\A_o\A_p^{-1/2}$ are functions of $(\Lambda, Q)$ jointly and
not of $\Lambda$ alone (the matrix also depends on $U$, which those invariants
do not see): with $\A_p = V\Lambda V^\top$,
$V = UQ$, it ranges from $I_d$ at $Q = I$ to the Haar average
$\E\operatorname{tr}\M = (\operatorname{tr}\Lambda)(\operatorname{tr}\Lambda^{-1})/d$
at random $Q$ (Proposition~\ref{prop:volume}), with $\det\M = 1$ throughout.
Matching the eigenvalue profile is therefore not sufficient (item (i)). The
flat probe is also undamaged against a steep private spectrum, but that
observation is not a corollary of the law: a flat probe does not share the
eigenvalue multiset, so $\M = \Lambda_o$ and $\det\M = \det\Lambda_o \neq 1$,
and the statement belongs to the empirical damage convention of
Section~\ref{sec:anatomy} rather than to
Proposition~\ref{prop:volume}. This is the precise content of the title.
\end{corollary}

\begin{proposition}[Whitened-volume law]
\label{prop:volume}
Let $\A_o = U\Lambda U^{\top}$ and $\A_p = V\Lambda V^{\top}$ be positive
definite with the same eigenvalue multiset
$\Lambda = \operatorname{diag}(\lambda_1 \ge \dots \ge \lambda_d > 0)$ and
eigenbases related by an orthogonal $Q$ ($V = UQ$), and write
$\M = \A_p^{-1/2}\A_o\A_p^{-1/2}$. Then:
\begin{enumerate}
\item[(i)] $\operatorname{tr}\M = \operatorname{tr}\!\left(\Lambda^{-1}
Q^{\top}\Lambda Q\right) = \sum_j \lambda_j \left(Q\Lambda^{-1}
Q^{\top}\right)_{jj}$ and $\det\M = 1$: misalignment spreads the residual
without biasing it.
\item[(ii)] Averaged over a Haar-random $Q$,
$\E\,\operatorname{tr}\M = \left(\operatorname{tr}\Lambda\right)
\left(\operatorname{tr}\Lambda^{-1}\right)/d \;\ge\; d$,
with equality if and only if $\Lambda$ is a multiple of the identity. The
white probe is therefore the unique spectrum whose residual is
basis-independent, and the expected damage of a matched non-flat spectrum
under a random basis is
$\kappa_{\mathrm{eff}}(\Lambda) = \left(\operatorname{tr}\Lambda\right)
\left(\operatorname{tr}\Lambda^{-1}\right)/d^{2} \ge 1$,
the ratio of the arithmetic mean of the spectrum to its harmonic mean.
\end{enumerate}
The proof is in \appref{app:proofs}; the identity case $Q = I$ gives
$\M = I_d$ exactly, and the flat case $\Lambda \propto I$ gives
$\M = I_d$ for every $Q$.
\end{proposition}

The volume scan of Observation~\ref{prop:basisfirst}(iii) rotates a
\emph{prefix} of the probe, not all of it, so it needs the partial version
of the law. The next two results supply it. The lemma reduces the damage of a
rotated prefix to a product of two partial sums, and the theorem shows that
this product grows by at least one unit per direction added, so the closed form
is monotone by construction and the measured scan monotone in practice.

\begin{lemma}[Partial-volume trace identity]
\label{lem:partial}
Let $Q$ be Haar-random on the top-$K$ eigendirections and act as the identity
on the orthogonal complement, with the expectation taken over $Q$, and put
$S_K = \sum_{i \le K}\lambda_i$ and
$T_K = \sum_{i \le K}\lambda_i^{-1}$. Then
$\E\,\operatorname{tr}\M = (d-K) + S_KT_K/K$. The rotated block enters only
through the product of its two partial sums, and at $K = d$ the identity
collapses to the full-volume law of
Proposition~\ref{prop:volume}(ii).
\end{lemma}

\begin{theorem}[Prefix growth is at least linear]
\label{thm:prefix}
Write $f(K) = S_KT_K/K$. For every $1 \le K < d$,
$f(K+1) \ge f(K) + 1$, with equality if and only if
$\lambda_1 = \dots = \lambda_{K+1}$. Hence
$\tau(K) = (f(K)-K)/d$ is nondecreasing in $K$, and $\tau \equiv 0$ on
$\{1,\dots,d\}$ if and only if the spectrum is flat. A flat spectrum is
therefore the unique spectrum with $\tau\equiv 0$, which is the prefix form of the
basis independence in Proposition~\ref{prop:volume}(ii).
\end{theorem}

Three readings follow. Proposition~\ref{prop:volume}(ii) explains the
white-noise reference: a flat spectrum is the only spectrum whose residual
does not depend on the basis, which is why every aligned-probe measurement in
Observation~\ref{prop:basisfirst} is quoted against it. Lemma~\ref{lem:partial}
turns the volume scan into a one-line formula, and
Theorem~\ref{thm:prefix} turns it into an ordering statement. The
misaligned volume fraction $\tau(K) = (f(K)-K)/d \ge 0$ is the excess of the
partial whitened volume over its uniform value; it is nondecreasing, it
vanishes on the whole range only for a flat spectrum, and the critical volume
is where it first crosses the damage threshold. For a power law
$\lambda_j \propto j^{-\gamma}$ the ratio $f(K)/K$ scales as
$\zeta(\gamma)K^{\gamma-1}/(\gamma+1)$ (Euler--Maclaurin,
\appref{app:proofs}), which places the measured ordering
$K^{\star} = 128, 128, 192, 320$ of Table~\ref{tab:volume} on a family indexed
by the steepness $\gamma$.

The closed form stops there, and the reason is not a missing constant.
Recomputed on the archived private spectra the absolute scale is disconnected
from the measured anisotropy (the flattest modality has
$f(512) = 1.6\times10^{3}$ with damage $0.98$), and over fifty probe-spectrum
points the matched idealization misses the measured anisotropy by a factor
between $0.8$ and $27.8$, with a non-monotone error in $K$. The next
proposition shows that no repair of the constant can succeed: the spectrum
fixes $S_K$ and $T_K$, but the damage also depends on how much of the oracle
eigenbasis the rotated prefix covers, so two oracles with the same eigenvalues
and different bases land on different curves.

\begin{proposition}[Eigenbasis overlap is not a function of the spectrum]
\label{prop:specnotenough}
Fix the probe spectrum $\Lambda$ and consider oracles whose spectra depart from
it by a relative $10^{-2}$, on $\lambda_i = i^{-1.8}$ with $d = 400$ and
$K = 100$. The anisotropy
$d\operatorname{tr}(\M^2)/\operatorname{tr}(\M)^2-1$ of the residual is then
not a function of the pair $(\Lambda_o,\Lambda_p)$. With a near-matched
eigenbasis it is $0$ on the unrotated arm and $105.5 \pm 10.6$ after rotating
the top $K$ directions over fifty Haar draws; with a Haar-incoherent
eigenbasis and the same eigenvalues the two arms read $126.4$ and $126.6$.
Spectra that agree to $1\%$ therefore produce anisotropies spanning from $0$
to $126$, so what sets the damage of a rotated prefix is the overlap between
the prefix and the oracle eigenbasis.
\end{proposition}

\subsection{What the probe family actually matches}

The $1/f^{\alpha}$ family works through two mechanisms that the spectral-shape
hypothesis conflates. First, its members inherit an approximately
Fourier-aligned basis, the first-order requirement on natural and medical
imagery. Second, within that basis the exponent tunes the activation-spectrum
steepness: the oracle exponent $\beta_o$ steepens from the first hidden layer
to the output on all five modalities (depth-4 fits of $-0.03$ to $-0.26$ per
layer), and on non-flat layers the layer-optimal probe is the one whose
activation exponent matches $\beta_o$, landing within $10\%$ of the empirical
optimum on $26$ of $31$ configurations at $\alpha \in [0,3]$ ($24$ of $43$ over
$[0,6]$), at zero tuned parameters (Figure~\ref{fig:beta}). Two
qualifications bound the rule. It consumes a private statistic: $\beta_o$ is a
private quantity, so the recipe is one-parameter and tuning-free but not fully
data-free. The support rate also depends on the grid: with a gain-$5\%$
criterion the matched exponent lands within $0.15$ of $\beta_o$ on $21$ of $43$
configurations (median gap $0.22$), so the rule is a trend, not an exact law.
And it fails at both spectral extremes, on flat input layers where no probe
resolves the bulk and on steep modalities the family cannot reach
($54.2\times$ on the OCTMNIST input layer).

The $1/f$ motivation survives in a narrower form. Self-organized criticality
explains the ubiquity of $1/f$ statistics in extended dynamical systems
\citep{bak1987}, and predictive coding assigns precision weights to prediction
errors in a hierarchy \citep{friston2010}; in our measurements $\alpha$ is a
tunable prior over spectral precision that helps where the trained bulk is
itself scale-free and hurts where it amplifies spikes. The mechanism is basis
alignment plus bounded precision matching, not spectral mimicry.

\subsection{Universality across datasets and architectures}
\label{sec:univ}

The verdict is not an artefact of the original five modalities: the separation
reproduces on seven of the eight configurations of Figure~\ref{fig:basis}
(Table~\ref{tab:univ}\ifLong\else{} in \appref{app:records}\fi). The exception
is PathMNIST, whose input spectrum is nearly isotropic
($bulk_1 \approx 0.98$ even for white noise), the $\Lambda \propto I$ case of
Proposition~\ref{prop:volume}. The resolution-and-channels control is sharp:
DermaMNIST at $28\times28$ grayscale and the same data at $64\times64$ RGB
agree to within $0.002$ on the aligned and white arms and to within $0.2\%$ on
the Haar arms ($bulk_1^{\mathrm{psdF}} = 0.968$ against $0.966$,
$bulk_1^{\mathrm{psdR}} = 86.89$ against $86.69$), so the law is a property
of the basis relation and not of the input representation.

A convolutional oracle behaves the same way: on two conv layers the Haar-basis
probe collapses onto the white baseline ($0.49$--$0.67$ against
$0.50$--$0.67$) while the Fourier-aligned probe sits at one quarter to one
sixth of it ($0.10$--$0.12$), and on the MLP oracle the Haar mismatch
amplifies to $89.8\times$ on DermaMNIST (Table~\ref{tab:univarch}).

\section{From operator residual to downstream utility}
\label{sec:downstream}

\begin{observation}[Dose-response]
\label{prop:dose}
For per-sample-clipped DP-SGD trained from scratch with a preconditioner
re-estimated each epoch from a fixed source, the final utility is
monotonically ordered by the source's bulk residual anisotropy, while the
missing spike mass does not predict utility. On PathMNIST over the source
family $\{$private, aligned-public, cross-domain public$\}$, the
bulk-only Spearman correlation with accuracy is $-1.0$ (Pearson $-0.878$
including all five sources), replicated over three seeds.
\end{observation}

Observation~\ref{prop:dose} is a family-level statement: ranking probe
sources by their bulk residual orders their downstream utility. A natural
extrapolation, already suggested by the layer-wise matching law, would be to
optimize each layer's probe exponent independently. We tested this
extrapolation directly with a five-arm design on three modalities
(private oracle; the oracle spectrum in a Haar basis; white; global exponent at
the layer-mean optimum; per-layer exponent), with balanced
accuracy as the primary metric, a mean-of-last-three-epochs statistic, and
forty seeds per arm (Table~\ref{tab:downstream}).

\begin{table}[!t]
\centering
\caption{Downstream accuracy under five probe regimes (mean of the last
three epochs of balanced accuracy, mean $\pm$ std over forty seeds,
$\sigma = 0$ DP-SGD, eight epochs). In this convention a constant predictor
scores $1/10$ and uniformly random labels $|C|/100$; the data-free arms clear
it by at most $0.012$, the oracle by $0.023$--$0.078$. \emph{eigR} is the
oracle spectrum in a Haar basis, from a run that reproduces the oracle and
white columns to $0.002$. The
oracle arm is an upper reference, not a method.}
\label{tab:downstream}
\small
\setlength{\tabcolsep}{3pt}
\begin{tabular}{lccccc}
\toprule
modality & oracle & eigR & white & global $\alpha^{\ast}$ & tailored \\
\midrule
DermaMNIST & $\mathbf{0.123}$ & $0.121$ & $0.104$ & $0.111$ & $0.108$ \\
           & $\pm.017$ & $\pm.009$ & $\pm.009$ & $\pm.023$ & $\pm.019$ \\
OCTMNIST   & $\mathbf{0.126}$ & $0.104$ & $0.101$ & $0.103$ & $0.103$ \\
           & $\pm.009$ & $\pm.002$ & $\pm.007$ & $\pm.007$ & $\pm.004$ \\
CIFAR-10   & $\mathbf{0.178}$ & $0.138$ & $0.110$ & $0.109$ & $0.107$ \\
           & $\pm.013$ & $\pm.012$ & $\pm.008$ & $\pm.012$ & $\pm.010$ \\
\bottomrule
\end{tabular}
\end{table}

Over forty seeds no data-free arm separates from
white: the largest gap is $+0.8\%$ (global exponent, DermaMNIST, $1.9$
standard errors), and the tailored arm stays within noise ($+0.4\%$, $+0.2\%$,
$-0.2\%$). The four-seed gap on
DermaMNIST ($+3.2\%$ at $3.6\sigma$) shrinks to $+0.8\%$ at forty seeds, as
the flat operator map between exponents $2.5$ and $6$ predicted. On
OCTMNIST the operator map's input-layer disaster zone ($54.2\times$ residual
range) produces no downstream catastrophe at all, and on CIFAR-10 the oracle climbs
to $0.178$ and the same spectrum in a Haar basis to $0.138$, both above white. Two conclusions follow. The residual orders sources
(Observation~\ref{prop:dose}) but not per-layer magnitudes: at the oracle
spectrum a Haar basis raises the input-layer bulk residual to $32$--$83$, the
psdR level, and costs $0.0$, $2.2$ and $3.9$ points of balanced accuracy
(Table~\ref{tab:downstream}). Against white the same spectrum in a random
basis keeps $+1.5$, $+0.2$ and $+2.9$: on DermaMNIST the gain is all spectral,
on OCTMNIST all alignment, on CIFAR-10 split. And the headroom of data-free
preconditioning is the observed oracle-minus-white gap, $+1.9\%$, $+2.4\%$ and
$+6.8\%$ over forty seeds ($2.39$ on OCTMNIST, unrounded), estimates on this
scale rather than bounds.

The size of the null is a property of the design: the smallest effect forty
seeds resolve ($80\%$ power, two-sided $5\%$) is $0.3$ to $1.1$ points across
the three modalities, against the largest data-free gap above, while the
oracle gap clears it on all three (Table~\ref{tab:power}). Ten seeds resolved
nothing below $1.7$ points on the noisiest modality, so the null is a
resolution limit and not an underpowered test.

Against the headroom rather than in absolute points the split is stark: every
data-free arm stays within $1.9$ standard errors of the flat probe, at most
$56\%$ of the oracle-minus-white gap where that gap is smallest and none of it
on the other two, while public-data curvature, which also costs no privacy
budget, clears the same bar on both (Table~\ref{tab:public}): an independent
five-arm replication returns $+0.92\pm0.50$ on the leading data-free arm,
TissueMNIST curvature matches the private oracle on DermaMNIST ($+1.69$ against
$+1.65$), and the more distant PathMNIST reaches $59\%$ of it on OCTMNIST. The dichotomy
predicts this: a public set can carry the data-induced spikes that no probe can
reconstruct, so wherever the two distributions overlap the public route
inherits the part of the problem that is recoverable in principle, while a
probe retains only the architecture-determined bulk. The practical reading is
that a data-free
probe is worth its zero cost when no comparable public data exists, and the
choice between the routes is a question about the data, not about the probe
spectrum.

\section{What the privacy mechanism destroys}
\label{sec:negative}

\subsection{The ordering survives below a noise cross-over}

The mechanism experiment below fixes the learning rate at $0.1$, so it cannot
separate ``the probe is wrong'' from ``the optimizer can no longer learn''. We
swept the noise multiplier against the learning rate over the same protocol
(120 runs\ifLong; Table~\ref{tab:exp21}\else; Figure~\ref{fig:crossover}\fi).
At the lower rate ($0.03$) a genuine ordering
exists below a cross-over: at $\sigma = 0.2$ ($\varepsilon = 274$) both matched
sources sit above the $0.143$ majority-class rate and the flat white probe
below it, and between $\sigma = 0.2$ and $\sigma = 0.3$ ($\varepsilon = 86$) the
ordering \emph{inverts}, the white probe overtaking both and staying ahead at
$\sigma = 0.5$. Two readings are
consistent and not separated here: the matched probe whitens a spike the noise
has removed, and its inverse square root amplifies noise in the
small-curvature directions a flat probe leaves alone.

\ifLong
\begin{table}[!t]
\centering
\caption{Noise tolerance and the source-ordering cross-over (PathMNIST, four
seeds, mean of the last three epochs, learning rate $0.03$; majority-class
rate $0.143$; $\varepsilon$ from RDP accounting). The matched sources lead
below $\sigma = 0.3$; the flat white probe leads above it.}
\label{tab:exp21}
\small
\setlength{\tabcolsep}{4pt}
\begin{tabular}{lcccc}
\toprule
$\sigma$ & $\varepsilon$ & private & aligned & white \\
\midrule
$0.0$ & --        & $0.322$ & $0.329$ & $0.130$ \\
$0.1$ & $1.1{\times}10^{4}$ & $0.217$ & $0.222$ & $0.132$ \\
$0.2$ & $274$     & $0.155$ & $0.157$ & $0.134$ \\
$0.3$ & $86$      & $0.126$ & $0.132$ & $\mathbf{0.155}$ \\
$0.5$ & $16.6$    & $0.110$ & $0.114$ & $\mathbf{0.128}$ \\
\bottomrule
\end{tabular}
\end{table}
\fi

\subsection{A mechanism experiment under real noise}

In a mechanism experiment (75 runs: 5 training sources $\times$ 5 noise
multipliers $\times$ 3 seeds, PathMNIST, the same MLP and DP-SGD protocol as
above with clip $1.0$, $q=0.0128$, $T=948$), RDP accounting gives
$\varepsilon = 16.6$ at $\sigma=0.5$ and $\varepsilon = 2.35$ at $\sigma=1$
($\delta = 5\times10^{-5}$, Table~\ref{tab:privacy}). At $\sigma=0$ the
aligned public probe matches the private oracle ($0.267$ against $0.264$ final
accuracy, majority-class rate $0.143$) while the two data-free probes stay near
chance ($0.119$ and $0.133$), the ordering of
Section~\ref{sec:downstream} on this modality.
At this fixed rate, for every $\sigma>0$ the three trained-source
arms fall below the majority-class rate ($0.098$--$0.105$), the private oracle
included, while the two synthetic arms bracket it ($0.086$--$0.153$);
there is no ordering to recover, which is why the cross-over sweep above varies
the learning rate.

The spectrum explains where the failure sits (Figure~\ref{fig:dp} in
\appref{app:dpacc}, Table~\ref{tab:exp19live}). On the true-DP models the top-1 fraction of the
first hidden activation factor drops from $0.330$ to $0.025$ the moment
noise is on ($13\times$) and then freezes, while the error-side factor first
flattens and then re-concentrates ($0.203 \to 0.030 \to 0.138$ across
$\sigma = 0, 0.5, 4$) and the activation tail exponent flattens from $-0.70$
to $-0.26$. Probe mismatch follows suit: the log-ratio dispersion of the
aligned public probe stays flat and an order of magnitude below every
alternative ($0.028$--$0.037$ across all $\sigma$), yet this structure is
operationally irrelevant: the bottleneck has moved from preconditioning to
learnability, since the preconditioner acts before the privacy mechanism and
cannot reshape isotropic noise after the fact.

\subsection{Measurement pitfalls}

Two pitfalls bound the theory (\appref{app:pitfalls}): single-epoch accuracy
inverts the probe-exponent order on DermaMNIST ($66.9\%$ majority class), so
downstream claims use balanced accuracy over the last three epochs; and the
dispersion diagnostic the shape hypothesis recommends anti-predicts the bulk
residual on $12$ of $43$ layer-configurations, so a diagnostic must be
validated against the quantity it proxies.

\section{Practical implications}
\label{sec:implications}

The account condenses to a default recipe at the operator level: one global
probe exponent per
modality, chosen once by matching the probe activation-spectrum steepness to
the trained bulk (a one-parameter, zero-tuning rule with $26/31$
operator-level support on the moderate grid); an aligned basis over spectral
mimicry; no per-layer customization, which does not transfer; and
data-induced spikes left to per-sample clipping. On the modalities tested the utility cost relative to the oracle is $2$ to $7$
points of balanced accuracy,
and the decomposition applies to any blockwise preconditioner that separates
an activation-side from an error-side factor.

\ifLong
\section{A data-free preconditioning protocol}
\label{sec:protocol}

The account condenses into the procedure of Algorithm~\ref{alg:recipe}: build
both factors from a Fourier-aligned probe, fix a single global exponent
(matching one private bulk statistic when one can be queried), precondition,
and leave data-induced spikes to per-sample clipping, which bounds the
influence of any single direction.

Complexity is far better than a dense Fisher inverse. For $L$ layers with
input widths $d^{\mathrm{in}}_l$ and output widths $d^{\mathrm{out}}_l$,
factor estimation costs
$\sum_l O(n_p d^{\mathrm{in}\,2}_l + n_p d^{\mathrm{out}\,2}_l)$
multiplications for the second-moment accumulations plus one probe
forward--backward pass; storage is
$O(\sum_l d^{\mathrm{in}\,2}_l + d^{\mathrm{out}\,2}_l)$; and each epoch
requires one symmetric eigendecomposition per factor. Inverting the layer's
full Fisher block would instead cost
$O\!\left((d^{\mathrm{in}}_l d^{\mathrm{out}}_l)^{3}\right)$. The optional
boundary check of Section~\ref{sec:pillarT2} is $O(d\log d)$: it reduces to
sorting the leading eigenvalues and comparing the leader with the bulk
scaling.

\begin{algorithm}[t]
\caption{Data-free KFAC preconditioning with spike-aware defaults}
\label{alg:recipe}
\begin{algorithmic}[1]
\REQUIRE trained model $f_\theta$; per-sample clip $C$; noise multiplier
$\sigma$; probe exponent $\alpha$ (or a conservative default); optional
private bulk statistic $\beta_o$.
\STATE Build the probe set $P$ from Fourier-aligned $1/f^{\alpha}$ colored
noise at unit variance, $|P| = n_p$ samples; keep the native basis, no basis
randomization.
\STATE For every layer $l$ estimate
$\A_l(P) = \E_P[a_l a_l^{\top}] + \varepsilon I$ and
$\G_l(P) = \E_P[\delta_l \delta_l^{\top}] + \varepsilon I$.
\STATE If $\beta_o$ is queryable, pick the $\alpha$ whose probe activation
exponent matches $\beta_o$; otherwise keep a single global $\alpha$ (do not
tailor per layer, Section~\ref{sec:downstream}).
\FOR{each epoch}
\STATE Re-estimate $\A_l(P)$, $\G_l(P)$ and form
$\mathsf{P}_l = \G_l(P)^{-1/2}\,\A_l(P)^{-1/2}$.
\STATE Per-sample-clip gradients at $C$, add Gaussian noise of multiplier
$\sigma$, precondition, and take a step.
\STATE Spike rule: leave the leading direction to clipping; do not sharpen it
beyond the bulk edge.
\ENDFOR
\STATE Optionally, and only where a valid bulk model is available, compare
the probe spectrum edge with the bulk edge (Lemma~\ref{lem:bbp}) to decide
whether any spike is recoverable.
\end{algorithmic}
\end{algorithm}
\fi

\section{Discussion}
\label{sec:discussion}

\textbf{What is and is not private.} Behind the privacy wall, the
architecture-determined bulk of $\A$ and the entire $\G$ factor are
reconstructible from probes: they are statements about the model class, not
about the data. What is not reconstructible is the detached data-induced
spike, whose threshold the BBP transition makes exact. In Kantian terms the
form that survives the wall is not a spectral shape but a basis, a restatement
of Corollary~\ref{cor:sufficient}: the reconstructible object is the pair
(spectrum, basis). Proposition~\ref{prop:volume} gives the closed form of that
distinction and Proposition~\ref{prop:specnotenough} shows that the basis
component cannot be recovered from the spectrum. This gap, not the probe
family, is the boundary of data-free preconditioning, and it links the account
to the spiked-covariance signal-propagation line \citep{wang2024}, where our
bulk and spike are the nonlinear analogues of the conjugate-kernel bulk and
signal eigenvalues.

\textbf{Falsifiable predictions.} (P1) A modality with a nearly isotropic
input covariance admits no data-free probe that reduces the residual; the
PathMNIST input spectrum is a witness, and any aligned probe that helps despite
$\Lambda \propto I$ would falsify it. (P2) The basis-first law depends only on
a blockwise preconditioner's activation factor, so it should transfer to the
coordinate-whitening preconditioners of the Shampoo and Muon families
\citep{zeta2026}, whose whitening plays the role of $\A^{-1/2}$ here. (P3)
Under DP noise the source ordering predicts utility only below the noise level
at which learning survives; Section~\ref{sec:negative} locates the collapse
side of this cross-over. \ifLong The protocol of Section~\ref{sec:protocol} is the
operative form of these predictions: fix the basis, choose the spectrum second,
and pre-check the detachment threshold from $d/n$ where a bulk model is
available. \else Operationally: fix the basis, then the spectrum, then check
the detachment threshold wherever a bulk model is available.\fi

\textbf{Relation to criticality accounts.} The $1/f$ statistics of natural
inputs have been read through self-organized criticality \citep{bak1987} and
through hierarchical precision weighting \citep{friston2010}. Our results
support a bounded version: a scale-free probe is a motivation space whose
exponent helps where the trained bulk is itself scale-free and hurts where it
amplifies data-induced spikes.

\textbf{Limitations.} All experiments use $28\times28$ grayscale inputs, a
$32\times32$ CIFAR-10 subset, and $64\times64$ RGB DermaMNIST; the
convolutional oracle uses an unfold approximation, and normalization and
attention layers are untested. Scale is modest (single $24$GB device); the
downstream comparison uses forty seeds, the dose-response family three and the
noise sweep four, so the reported gaps are estimates, not bounds. The noise
cross-over (Section~\ref{sec:negative})
sits at weak privacy in this small-model protocol, where learning survives only
for $\varepsilon$ well above reporting thresholds; widening the set
$4.5\times$ to $\varepsilon = 9.99$ at $\sigma = 0.5$ still leaves the three
real-source arms at $0.125$--$0.130$ against $0.143$. Theory-side, the ordering of $K^{\star}$ is monotone in the
spectrum (Lemma~\ref{lem:partial}, Theorem~\ref{thm:prefix}), and that is all
the spectrum gives (\S\ref{sec:anatomy}): the matched-spectrum idealization
misses the measured anisotropy by $0.8$--$27.8$, and quantifying the eigenbasis
overlap that sets it is open. On the probe side, the exponent-matching rule
consumes one private statistic (within $0.15$ on $21$ of $43$ configurations),
so a public proxy and an exact rule are open.

\ifLong\else
\bibliography{refs}
\bibliographystyle{IEEEtranN}

\appendix
\fi

\section{Overview of the account}
\label{app:overview}

\begin{figure*}[!t]
\centering
\includegraphics[width=0.98\textwidth]{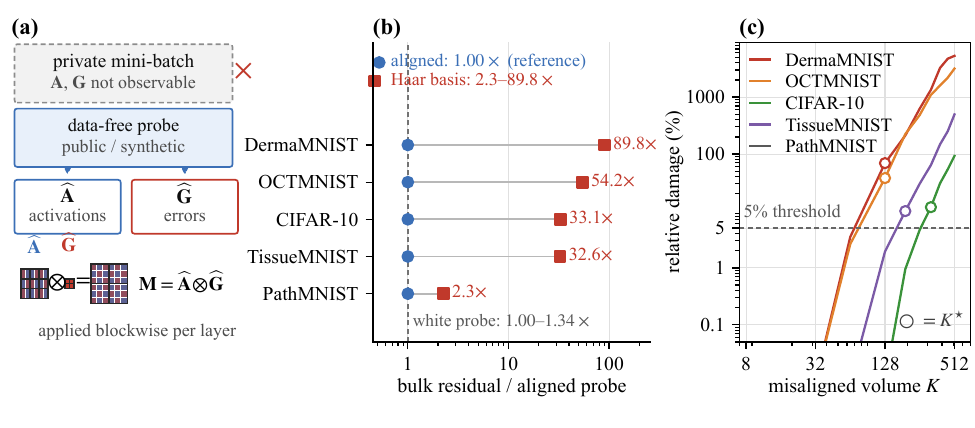}
\caption{Overview. (a)~Data-free two-factor anatomy: the private
mini-batch is unavailable, so a probe built from public or synthetic
spectra supplies both Kronecker factors, combined into the preconditioner
$\mathbf{M}=\widehat{\mathbf{A}}\otimes\widehat{\mathbf{G}}$ applied
blockwise per layer. (b)~The basis-first inversion (input layer, eight
dataset configurations): relative to the white probe, a probe with the
matched spectrum in the aligned basis sits at $0.75$--$1.00\times$ on
every modality, while the same eigenvalues in a Haar-random basis inflate
the bulk residual by $2.3\times$ (PathMNIST) to $89.8\times$
(DermaMNIST); the two arms share a bitwise-identical log-ratio dispersion
($0.286130$ on DermaMNIST). (c)~The whitened-volume law: relative
bulk-residual damage against the number $K$ of rotated top
eigendirections; open circles mark the empirical $K^{\star}$ ($128$ on
OCTMNIST and DermaMNIST, $192$ on TissueMNIST, $320$ on CIFAR-10, not
reached on PathMNIST up to $K=512$). The closed forms
$\det \mathbf{M}=1$ and
$\mathbb{E}[\operatorname{tr}\mathbf{M}]
=(\operatorname{tr}\Lambda)(\operatorname{tr}\Lambda^{-1})/d$ place this
ordering on a family indexed by spectral steepness; the magnitude is not a
function of the spectrum
(Proposition~\ref{prop:specnotenough}).}
\label{fig:concept}
\end{figure*}

Figure~\ref{fig:concept} summarizes the setting, the basis-first
inversion, and the volume law in one panel.

\begin{figure*}[!t]
\centering
\includegraphics[width=0.93\textwidth]{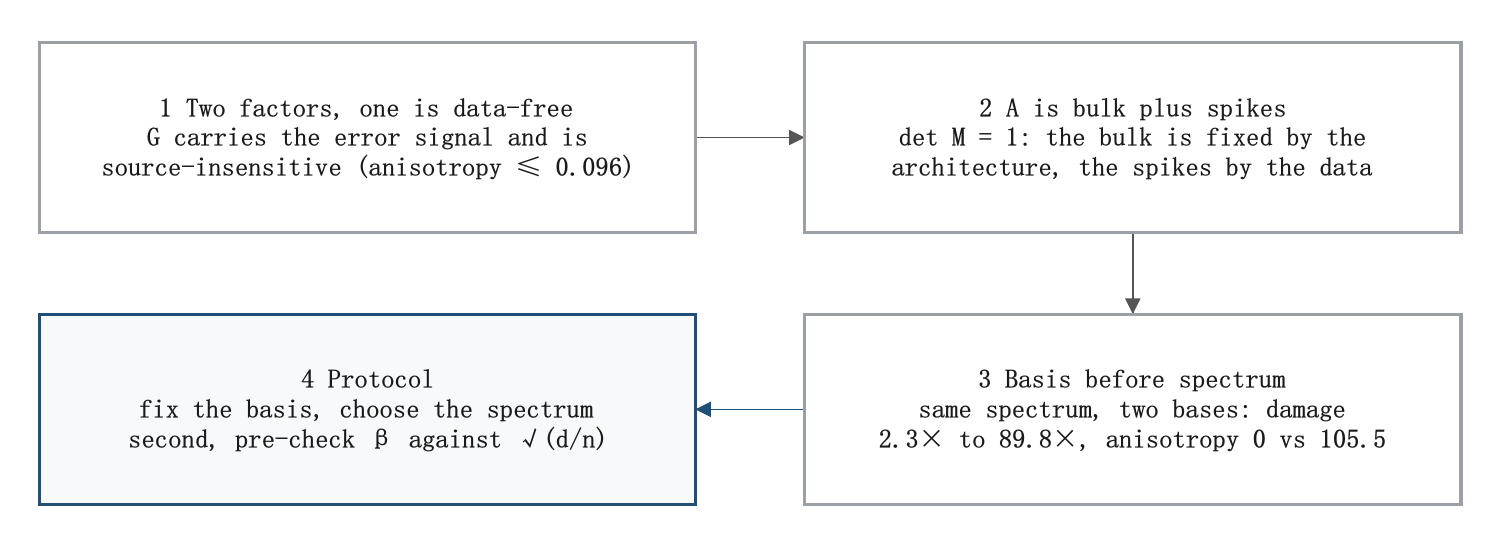}
\caption{The account as a decision cascade. Step~1 splits the two factors,
step~2 splits the activation factor into an architecture-determined bulk and
data-induced spikes, step~3 is the impossibility: two bases at one spectrum
separate by $2.3\times$ to $89.8\times$ while the spectra stay identical
(the measured separation of Observation~\ref{prop:basisfirst}(i), whose
impossibility proof is Proposition~\ref{prop:specnotenough}), and step~4 is the
order of decisions
that follows (\ifLong Section~\ref{sec:protocol}\else
Section~\ref{sec:implications}\fi). Each step carries the decisive
measurement rather than an illustration.}
\label{fig:cascade}
\end{figure*}

\section{Proofs}
\label{app:proofs}

\subsection{Proof of Proposition~\ref{prop:g}}

For (i), convexity of the operator norm and
$\|\delta\delta^{\top}\|\op = \|\delta\|^2$ give
$\|\E_S[\delta\delta^{\top}] - \E_{S'}[\delta'\delta'^{\top}]\|\op
\le \E_S\|\delta\|^2 + \E_{S'}\|\delta'\|^2$.
For (ii), at the last layer with cross-entropy loss
$\delta = p - e_y$ with $p$ on the simplex, so
$\|p - e_y\|^2 = \sum_i p_i^2 - 2p_y + 1 \le 2(1 - p_y)$ because
$\sum_i p_i^2 \le 1$. Taking expectations on both sources and applying (i)
gives $\|\G_l(S) - \G_l(S')\|\op \le 4\eta$ when both sources have mean
true-class confidence at least $1-\eta$; the two $\varepsilon I$ terms are the
same constant and cancel, which is why no additive noise term appears.
Part (iii) names what the bound does and does not buy: at initialization
$p(x)$ is uniform, so $\eta \approx 1 - 1/C$ and it is vacuous, and over the
measured range $\eta \in [0.655, 0.904]$ it stays one to nearly three orders
of magnitude above the gap it bounds, so it certifies that no modality term
enters without pinning the magnitude.

\subsection{Proof of Proposition~\ref{prop:t2}}

With $\A_o = (1-s^2)I_d + s^2 uu^{\top}$ and a spike-blind probe
$\A_p = (1-s^2)I_d$, we have
$\A_p^{-1/2} = (1-s^2)^{-1/2}I_d$ and therefore
$\M = \A_o/(1-s^2) = I_d + \frac{s^2}{1-s^2}uu^{\top}$.
The eigenvalues of $\M$ are $1 + s^2/(1-s^2) = 1/(1-s^2)$ in the direction
$u$ and $1$ on $u^{\perp}$, so the top eigenvalue is $1/(1-s^2)$,
monotone increasing in $s$ and independent of the basis. Part (ii) is
Lemma~\ref{lem:bbp} restated: the sample top eigenvalue of $\A$ detaches
from the Marcenko--Pastur edge if and only if $s^2/(1-s^2) > \sqrt{d/n}$, so below
the threshold the spike is statistically indistinguishable from the bulk
while above it the sample spike carries an isolated eigendirection.

\subsection{Proof of Lemma~\ref{lem:bbp} and Proposition~\ref{prop:t2bulk}}

For the lemma, divide $\Sigma = (1-s^2)I_d + s^2 uu^\top$ by $1-s^2$. The
sample covariance of the rescaled data is that of a rank-one spike of strength
$\beta = s^2/(1-s^2)$ over a unit isotropic bulk; the Baik--Ben
Arous--P\'ech\'e threshold for this model is $\beta > \sqrt{d/n}$
\citep{baik2005}, which gives $s^2/(1-s^2) > \sqrt{d/n}$ and
$s_c = (\sqrt{c}/(1+\sqrt{c}))^{1/2}$. The leading-order form $s_c = c^{1/4}$
is the $s \to 0$ limit and overstates the threshold by $0.110$ at
$c = 0.191$.

For the proposition, write $\Sigma = \Sigma_b + \beta uu^\top$ and let $m_b$ be
the Stieltjes transform of $\mu_b$. By the low-rank perturbation theorem of
\citet{bg2011}, the sample top eigenvalue admits an outlier above the bulk
edge if and only if $\beta$ exceeds the value at which the free-additive
fixed point for the outlier loses its solution above the edge; the
scaled-isotropic bulk $\Sigma_b = \sigma_b^2 I$ has an explicit transform
giving $\beta_c = \sigma_b^2 \sqrt{c}$, which reduces to the lemma. The numbers
quoted for the power-law bulks ($\beta_c/b_+ \approx 2.21$--$2.24$) come from the
crossing-based numerical study of Figure~\ref{fig:boundary}c and are a
numerical illustration rather than a closed form; they show that the boundary
depends on the bulk spectrum and not on its edge alone.

\ifLong\else
\subsection{Boundary refinement}

\begin{figure*}[!t]
\centering
\includegraphics[width=0.94\textwidth]{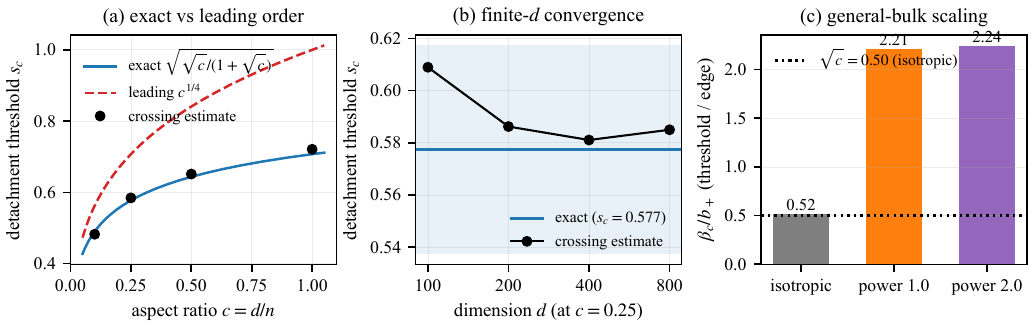}
\caption{Refinement of the failure boundary. (a) The exact threshold
$s_c = (\sqrt{c}/(1+\sqrt{c}))^{1/2}$ tracks the crossing-based empirical
estimate across aspect ratios, while the leading-order form $c^{1/4}$
overstates it. (b) The empirical threshold converges to the exact value as the
dimension grows at fixed $c = 0.25$. (c) For an anisotropic bulk the threshold
is a functional of the bulk spectrum: $\beta_c/b_+ \approx 0.52$ for an
isotropic bulk (with $\sqrt{c} = 0.5$, $b_+$ the population edge) and
$\approx 2.21$--$2.24$ for power-law bulks of exponent $1$ and $2$.
Numbers from verify\_boundary\_law (MC, archived
in figs/boundary\_refine\_data.json).}
\label{fig:boundary}
\end{figure*}
\fi

\subsection{Proof of Proposition~\ref{prop:volume}}

Write $D = \Lambda$ for the shared diagonal spectrum. Since
$\A_p = VDV^{\top}$ with $V = UQ$, cyclicity of the trace gives
$\operatorname{tr}\M = \operatorname{tr}(\A_p^{-1}\A_o)
= \operatorname{tr}(V^{-1}D^{-1}V^{-\top}UDU^{\top})
= \operatorname{tr}(D^{-1}Q^{\top}DQ)
= \sum_j \lambda_j (QD^{-1}Q^{\top})_{jj}$,
because $V^{\top}U = Q$. The determinant gives
$\det\M = \det\A_o/\det\A_p = 1$. This proves (i); the case $Q = I$ gives
$\M = D^{-1/2}DD^{-1/2} = I_d$.

For (ii), the diagonal of a Haar conjugate is exchangeable and sums to
$\operatorname{tr}D^{-1}$, so
$\E[(QD^{-1}Q^{\top})_{jj}] = \operatorname{tr}(D^{-1})/d$ for every $j$, and
linearity of the trace gives
$\E\,\operatorname{tr}\M = (\operatorname{tr}D)(\operatorname{tr}D^{-1})/d$.
Cauchy--Schwarz gives $(\operatorname{tr}D)(\operatorname{tr}D^{-1}) \ge d^2$
with equality if and only if all $\lambda_j$ are equal, which is (ii); the
flat case $\Lambda \propto I$ makes $\M = I_d$ for every orthogonal $Q$.

\subsection{Proof of Lemma~\ref{lem:partial}}
\label{app:partial}

Work in the probe eigenbasis, where $\A_p = VDV^{\top}$ and $V^{\top}U = Q$
with $Q = \operatorname{diag}(H, I_{d-K})$ for a $K\times K$ Haar block $H$
(the prefix is a permutation of coordinates, so no generality is lost). Then
$Q^{\top}DQ = \operatorname{diag}(H^{\top}D_KH, D_{S^{\perp}})$ and
\[
D^{-1}Q^{\top}DQ
= \begin{pmatrix} D_K^{-1}H^{\top}D_KH & 0 \\ 0 & I_{d-K}\end{pmatrix},
\]
so the trace separates into the rotated block and the untouched tail. For the
block, the $(i,i)$ entry is
$\lambda_i^{-1}\sum_{j \le K}\lambda_jH_{ji}^2$, hence
$\operatorname{tr}(D_K^{-1}H^{\top}D_KH)
= \sum_{i,j \le K}\lambda_i^{-1}\lambda_jH_{ji}^2$. A Haar matrix has
exchangeable rows of unit norm, so $\E[H_{ji}^2] = 1/K$ for every entry, and
linearity gives
\begin{align*}
\E\operatorname{tr}\M &= (d-K) + \frac{S_KT_K}{K}, \\
S_KT_K &= \textstyle\sum_{j \le K}\lambda_j \sum_{i \le K}\lambda_i^{-1}.
\end{align*}
At $K = d$ the tail vanishes and the identity returns
$S_dT_d/d = (\operatorname{tr}\Lambda)(\operatorname{tr}\Lambda^{-1})/d$,
the full-volume law.

\subsection{Proof of Theorem~\ref{thm:prefix}}
\label{app:prefix}

Let $\lambda = \lambda_{K+1}$, $S = S_K$ and $T = T_K$. Since the spectrum is
non-increasing, $\lambda_i \ge \lambda \ge \lambda_j$ for all $i,j \le K$, so
each of the $K^2$ pairs satisfies
\[
\frac{\lambda_i}{\lambda} + \frac{\lambda}{\lambda_j}
- \frac{\lambda_i}{\lambda_j} - 1
= \frac{(\lambda_i-\lambda)(\lambda_j-\lambda)}{\lambda\lambda_j} \ge 0 .
\]
Summing over the pairs gives $S/\lambda + \lambda T \ge f(K) + K$. Second,
$f(K+1) = (S+\lambda)(T+1/\lambda)/(K+1)$, and cancelling the common factor $K$
in the difference gives
\[
f(K+1) - f(K)
= \frac{S/\lambda + \lambda T + 1 - f(K)}{K+1} \ge 1 ,
\]
where the last step uses the pairwise inequality of the previous display.
The first inequality is an equality if and only if every pair has
$\lambda_i = \lambda$ or $\lambda = \lambda_j$, that is, if and only if
$\lambda_1 = \dots = \lambda_{K+1}$. In that case $f(K) = K$ and
$\tau(K+1) = \tau(K) = 0$; any strict inequality gives
$\tau(K+1) > \tau(K)$. Applying the dichotomy at every $K$ shows that
$\tau \equiv 0$ if and only if the whole spectrum is flat.

For the power law $\lambda_j = cj^{-\gamma}$ we have
$S_K = cH_K^{(\gamma)}$ and $T_K = c^{-1}H_K^{(-\gamma)}$, so
$f(K)/K = H_K^{(\gamma)}H_K^{(-\gamma)}/K^2$ does not depend on $c$.
Euler--Maclaurin gives
$H_K^{(\gamma)} = \zeta(\gamma) + O(K^{1-\gamma})$ for $\gamma > 1$ and
$H_K^{(-\gamma)} = K^{\gamma+1}/(\gamma+1) + O(K^{\gamma})$, hence
\[
\frac{f(K)}{K} = \frac{\zeta(\gamma)K^{\gamma-1}}{\gamma+1}
\big(1 + O(K^{-1}) + O(K^{1-\gamma})\big).
\]
The right side grows with $\gamma$ at fixed $K$ in the steep regime, which is
the direction used in the text; the monotonicity in $\gamma$ is verified
numerically over $\gamma \in (1,3)$ and is quoted as an asymptotic scaling,
not as part of the theorem.

\begin{figure*}[!t]
\centering
\includegraphics[width=0.95\textwidth]{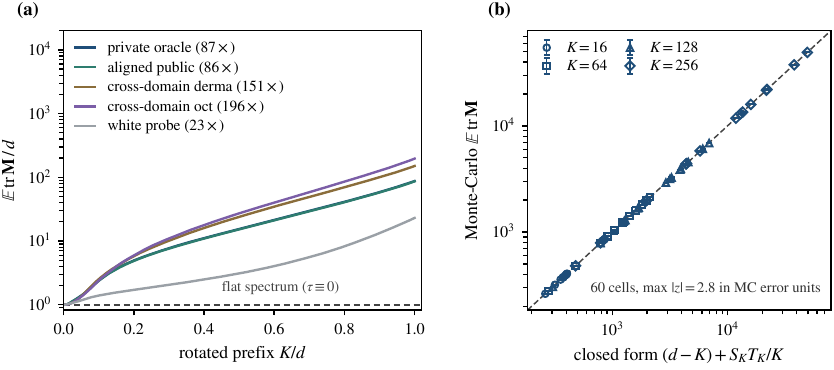}
\caption{The volume scan on measured activation spectra (PathMNIST, layer
$4$, $d = 257$, the five trained probe sources of
Section~\ref{sec:anatomy}). (a)~The prefix anisotropy
$\tau(K) = (f(K)-K)/d$ of Theorem~\ref{thm:prefix}, drawn as the
multiplicative trace factor $\E\operatorname{tr}\M/d = 1+\tau(K)$ that
Lemma~\ref{lem:partial} supplies. A flat spectrum stays at $1$ for every
$K$; every measured spectrum rises monotonically, and the legend carries
the full-prefix factor $1+\tau(d)$, which spans $23\times$ to
$196\times$ across the five sources. (b)~Monte-Carlo estimate of
$\E\operatorname{tr}\M$ under a Haar-rotated top-$K$ prefix against the
closed form $(d-K)+S_KT_K/K$ (five sources, three layers,
$K \in \{16,64,128,256\}$, fixed seed); every cell lies within the
Monte-Carlo error bars drawn.}
\label{fig:volscan}
\end{figure*}

\subsection{Construction for Proposition~\ref{prop:specnotenough}}
\label{app:specnotenough}

Take $d = 400$, $K = 100$, $\lambda_i = i^{-1.8}$, $\epsilon = 10^{-2}$, and
measure anisotropy as $d\operatorname{tr}(\M^2)/\operatorname{tr}(\M)^2 - 1$,
which is scale free and does not involve the clipped square root used in the
experiment code. Both oracles share the probe spectrum $\Lambda$ and depart
from it only by the factor $1+\epsilon$ on the eigenvalues.

\emph{Near-matched basis.} Let $\A_o = (1+\epsilon)U\Lambda U^{\top}$ with
$U = V$, the probe's own basis. Then
$\M_0 = \Lambda^{-1/2}(1+\epsilon)\Lambda\Lambda^{-1/2} = (1+\epsilon)I$ and
the aligned anisotropy is $0$ to machine precision
($2.2\times10^{-16}$). Rotating the top-$K$ probe directions by a Haar block
and averaging the anisotropy of $\M_K$ over fifty draws gives
$105.5 \pm 10.6$, so a rotation that leaves the spectrum untouched moves the
anisotropy from exactly $0$ to $105.5$.

\emph{Haar-incoherent basis.} Let $\A_o = (1+\epsilon)U\Lambda U^{\top}$ with
$U$ an independent Haar basis. The aligned arm reads $126.4$ and the rotated
arm $126.6 \pm 0.2$.

The two oracles have probe spectra that agree to $1\%$, and across their four
arms the anisotropy takes values from $0$ to $126$. Letting $\epsilon \to 0$
makes the spectral inputs coincide while those values persist, so no
functional of $(\Lambda_o,\Lambda_p)$ reproduces them. The mechanism is
the one the statement names: the rotated prefix adds $(d-K) + S_KT_K/K$ to the
trace, but the anisotropy of $\M$ is dominated by the overlap between the
oracle and probe bases, which the spectra do not encode. The check is
reproduced by \texttt{scripts/verify\_prop5.py} (fifty draws, seed fixed) and
archived in \texttt{docs/prop5\_check.json}.

\section{Experimental records}
\label{app:records}

\subsection{Metric conventions}

Four quantities recur in the tables and are defined once here.
\emph{Log-ratio dispersion} is the standard deviation of
$\log(\lambda^{o}_k/\lambda^{p}_k)$ over a common leading prefix of the paired
eigenvalues of the oracle and the probe, with each spectrum floored at
$10^{-10}$; it is zero only when the two profiles are proportional on that
prefix. \emph{Tail exponent} $\beta$ is the least-squares slope of
$\log\lambda_k$ against $\log k$ over the same prefix, so a steeper decay is a
more negative exponent. \emph{Fourier concentration} is the mean, over the
leading $32$ probe eigenvectors, of the largest squared coefficient of that
eigenvector in the two-dimensional discrete Fourier basis of the input grid; it
equals $1$ for a pure Fourier mode and falls to the Haar level
($\approx 8.5\times10^{-3}$) for a random basis. The \emph{relative damage} that
locates $K^{\star}$ in the volume scan is the bulk residual of the misaligned
top-$K$ arm divided by the bulk residual of the fully aligned arm on the same
layer, so $K^{\star}$ is the first $K$ at which that ratio exceeds $1.05$.

\subsection{Privacy budget}

RDP accounting (opacus $1.4.0$) for the mechanism experiment of
Section~\ref{sec:negative}: sampling probability $q = 256/20000 = 0.0128$,
$T = 948$ steps, $\delta = 5\times10^{-5}$. The $\sigma = 0$ row carries no
formal guarantee.

\begin{table}[!t]
\centering
\caption{Privacy budget of the DP-noise mechanism experiment
(PathMNIST, $n = 20000$, batch $256$, $T = 948$, $\delta = 5\times10^{-5}$).}
\label{tab:privacy}
\small
\begin{tabular}{cccc}
\toprule
$\sigma$ & $\varepsilon$ (RDP) & optimal order & steps \\
\midrule
$0.0$ & no formal guarantee & n/a & $948$ \\
$0.5$ & $16.556$ & $1.9$ & $948$ \\
$1.0$ & $2.350$ & $6.6$ & $948$ \\
$2.0$ & $0.780$ & $18.0$ & $948$ \\
$4.0$ & $0.338$ & $36.0$ & $948$ \\
\bottomrule
\end{tabular}
\end{table}

\subsection{Live spectral statistics under DP noise}

Means over $3$ seeds $\times$ $4$ epoch snapshots on the true-DP
(private-oracle) models; the probe-mismatch dispersion is measured on the
first hidden layer against the aligned public probe.

\begin{table}[!t]
\centering
\caption{Live-oracle spectral statistics of
Section~\ref{sec:negative} (private-oracle source, PathMNIST).
$\beta_o$ and $\beta_g$ are the live tail exponents of the activation and
error factors; top-1 fractions are spike shares of the live spectrum.}
\label{tab:exp19live}
\small
\setlength{\tabcolsep}{4pt}
\begin{tabular}{ccccc}
\toprule
$\sigma$ & top-1 ($A$, L2) & top-1 ($G$) & $\beta_o$ & $\beta_g$ \\
\midrule
$0.0$ & $0.330 \pm .018$ & $0.203 \pm .042$ & $-0.695$ & $-0.746$ \\
$0.5$ & $0.025 \pm .001$ & $0.030 \pm .006$ & $-0.265$ & $-0.561$ \\
$1.0$ & $0.020 \pm .001$ & $0.038 \pm .008$ & $-0.264$ & $-0.737$ \\
$2.0$ & $0.019 \pm .001$ & $0.093 \pm .043$ & $-0.264$ & $-1.057$ \\
$4.0$ & $0.018 \pm .001$ & $0.138 \pm .067$ & $-0.263$ & $-1.632$ \\
\midrule
\multicolumn{5}{l}{aligned-probe $\mathrm{lr\_std}$ (L2), per-$\sigma$ mean: $0.028$--$0.037$} \\
\bottomrule
\end{tabular}
\end{table}

\subsection{Accuracy under DP noise}
\label{app:dpacc}

\begin{figure*}[!t]
\centering
\includegraphics[width=0.96\textwidth]{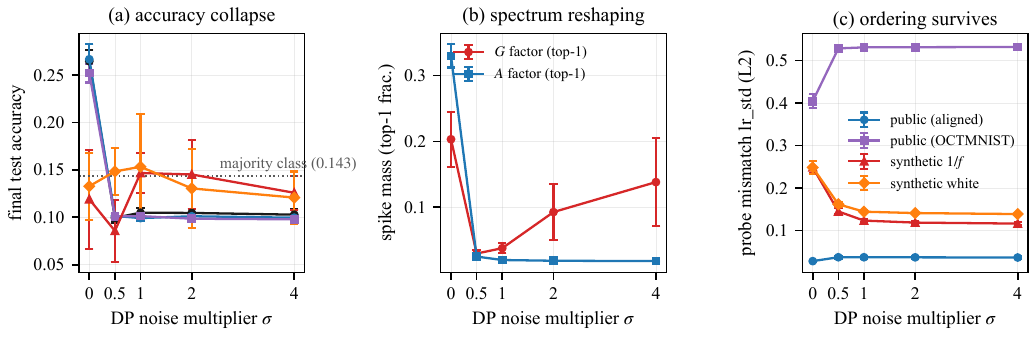}
\caption{DP-noise mechanism experiment (75 runs). (a) Once $\sigma>0$ the
final test accuracy of the three trained-source arms, private oracle
included, sits at or below the majority-class rate, while the two synthetic
arms stay at chance. (b) On the true-DP models, noise first removes the
activation-factor spike (top-1 fraction $0.330 \to 0.025$) and later
re-concentrates the error factor ($0.030 \to 0.138$). (c) The log-ratio
mismatch of the aligned public probe stays an order of magnitude below all
alternatives at every $\sigma$, but the surviving structure is
operationally irrelevant.}
\label{fig:dp}
\end{figure*}

Final test accuracy (mean over $3$ seeds) for the five training sources of
the mechanism experiment; the majority-class rate on this split is $0.143$.

\begin{table}[!t]
\centering
\caption{Final test accuracy by source and noise multiplier
(Section~\ref{sec:negative}; majority-class rate $0.143$).}
\label{tab:exp19acc}
\small
\setlength{\tabcolsep}{4pt}
\begin{tabular}{lccccc}
\toprule
source & $\sigma=0$ & $0.5$ & $1.0$ & $2.0$ & $4.0$ \\
\midrule
private oracle   & $0.264$ & $0.099$ & $0.105$ & $0.105$ & $0.103$ \\
public (aligned) & $0.267$ & $0.101$ & $0.099$ & $0.101$ & $0.100$ \\
public (OCTMNIST)& $0.252$ & $0.101$ & $0.101$ & $0.098$ & $0.098$ \\
synthetic $1/f$  & $0.119$ & $0.086$ & $0.147$ & $0.145$ & $0.126$ \\
synthetic white  & $0.133$ & $0.148$ & $0.153$ & $0.130$ & $0.121$ \\
\bottomrule
\end{tabular}
\end{table}

\subsection{Calibration of the error-energy bound}

Full record for the trajectory calibration of
Section~\ref{sec:calib}: $72$ source--layer--epoch configurations,
$3$ sources $\times$ $4$ layers $\times$ $6$ epochs.

\begin{table}[!t]
\centering
\caption{Error-energy bound calibration (exp.\ trajectory record).}
\label{tab:calib}
\small
\begin{tabular}{lc}
\toprule
quantity & value \\
\midrule
configurations & $72$ \\
energy-bound violations & $0$ \\
$\eta$-bound violations & $0$ \\
max ratio to (i) bound & $6.4\%$ \\
median ratio to (i) bound & $0.15\%$ \\
$\eta$ at epoch $0$ / $16$ & $0.904$ / $0.655$ \\
fixed-point discrepancy across sources & $2.6\%$ \\
\bottomrule
\end{tabular}
\end{table}

\subsection{Activation-spectrum matching and downstream transfer}

Figures behind the layer-optimal exponent rule of
Section~\ref{sec:anatomy} and the downstream experiment of
Section~\ref{sec:downstream}.

\begin{figure}[!t]
\centering
\includegraphics[width=0.82\columnwidth]{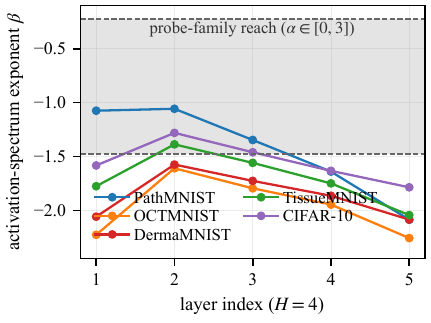}
\caption{Activation-spectrum matching: the oracle tail exponent $\beta_o$
steepens from the first hidden layer to the output layer on every modality,
while the reachable range of the $\alpha \in [0,3]$ probe family is a narrow
band. The layer-optimal probe exponent tracks the oracle exponent only where
the two ranges overlap.}
\label{fig:beta}
\end{figure}

\begin{figure*}[!t]
\centering
\includegraphics[width=0.64\textwidth]{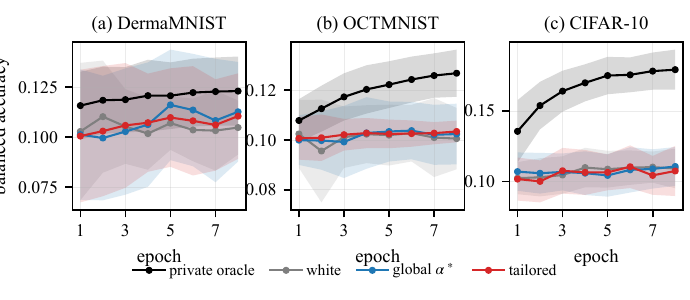}\\[10pt]
\includegraphics[width=0.65\textwidth]{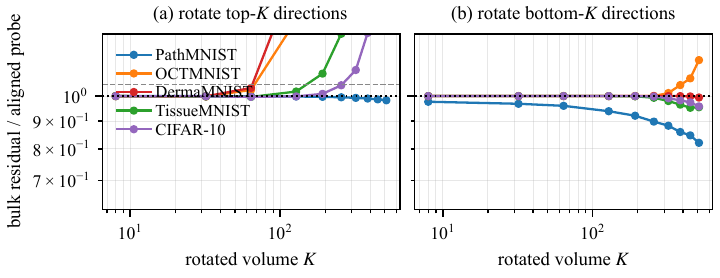}
\caption{Downstream transfer and the volume scan, both under the forty-seed
replication. \textbf{(a)}~Balanced accuracy per epoch (mean and standard
deviation over forty seeds) for the four probe regimes on DermaMNIST, OCTMNIST
and CIFAR-10. The global exponent leads on DermaMNIST by $+0.8\%$ ($1.9$
standard errors) and the per-layer tailored arm is within noise of
both white and global everywhere, so per-layer tailoring does not improve on a
single global exponent. \textbf{(b)}~Relative bulk residual against the number
$K$ of rotated top eigendirections; the critical volume moves monotonically
with the steepness of the data spectrum, and the bottom-block arm stays mild
everywhere.}
\label{fig:records}
\end{figure*}

\subsection{Rotation sweep}

\begin{figure*}[!t]
\centering
\includegraphics[width=0.87\textwidth]{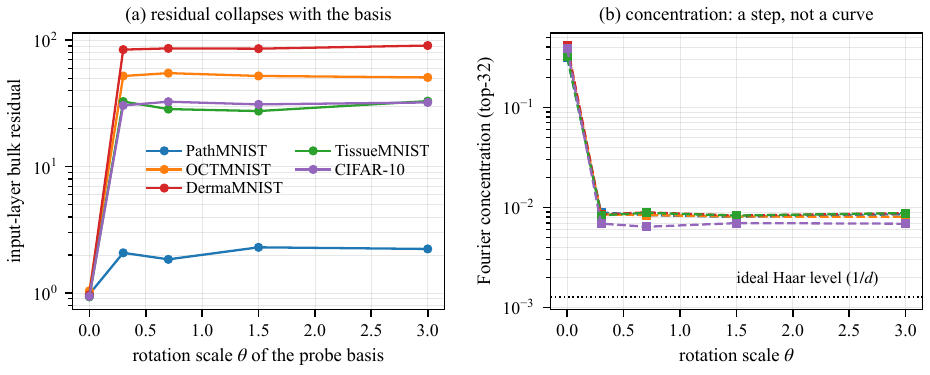}
\caption{Controlled verdict of the basis hypothesis. Left: bulk residual
against the rotation scale $\theta$ of the probe basis at fixed spectrum:
the residual is a step function of basis mixing, while the spectral
dispersion (right axis family) is bitwise invariant along the rotation.
Right: the loss of Fourier concentration of the probe; the random-basis
value is $\approx 8.5 \times 10^{-3}$.}
\label{fig:rotation}
\end{figure*}

Figure~\ref{fig:rotation} records the controlled rotation behind
Observation~\ref{prop:basisfirst}(ii): at a fixed spectrum, the bulk residual
is a step function of basis mixing while the spectral dispersion is bitwise
invariant.

\subsection{Public-data curvature against the data-free arms}

Table~\ref{tab:public} holds the paired differences of a five-arm replication
independent of Section~\ref{sec:downstream}, run to check that the null is not
an artifact of one seed block: both public arms separate from the flat probe on
both modalities at forty seeds, the nearer public set matches the private
oracle on DermaMNIST ($+1.69$ against $+1.65$) and the two public arms reach
$46\%$ and $59\%$ of it on OCTMNIST ($+1.07$ and $+1.38$ against $+2.33$).

\begin{table}[!t]
\centering
\caption{Public-data curvature against the data-free arms, forty seeds, paired
against the flat probe over the last three epochs, in balanced-accuracy points.
The public route consumes no privacy budget but changes the distribution; the
two public sets differ in their distance from the target modality. These rows
come from a dedicated five-arm replication that re-trains its own flat probe,
so its oracle and exponent differences sit $0.09$ to $0.23$ points away from
the corresponding entries of Table~\ref{tab:power}, within $1.3$ standard
errors of the paired difference; each table is paired within its own run.}
\label{tab:public}
\small
\setlength{\tabcolsep}{4pt}
\begin{tabular}{lrrrr}
\toprule
& \multicolumn{2}{c}{DermaMNIST} & \multicolumn{2}{c}{OCTMNIST} \\
\cmidrule(lr){2-3}\cmidrule(lr){4-5}
arm & $\Delta$ & s.e. & $\Delta$ & s.e. \\
\midrule
oracle (private)     & $+1.65$ & $0.27$ & $+2.33$ & $0.21$ \\
public (TissueMNIST) & $+1.69$ & $0.25$ & $+1.07$ & $0.13$ \\
public (PathMNIST)   & $+0.87$ & $0.22$ & $+1.38$ & $0.13$ \\
global exponent      & $+0.92$ & $0.50$ & $-0.10$ & $0.14$ \\
\bottomrule
\end{tabular}
\end{table}

\subsection{Resolution of the downstream design}

Table~\ref{tab:power} records the paired differences behind the null result of
Section~\ref{sec:downstream}: the smallest effect the forty-seed design
resolves on each modality, and the seed count a one-point effect would require.

\begin{table}[!t]
\centering
\caption{Resolution of the downstream design over forty seeds. Paired
difference against the flat probe over the last three epochs, its standard
error, the smallest effect resolvable ($80\%$ power, two-sided $5\%$), the
$95\%$ interval, and the seed count a one-point effect would need. Values are
balanced-accuracy points.}
\label{tab:power}
\small
\setlength{\tabcolsep}{3pt}
\begin{tabular}{llrrrr}
\toprule
dataset & arm & $\Delta$ & s.e. & MDE & $n_{1\%}$ \\
\midrule
DermaMNIST & oracle   & $+1.88$ & $0.32$ & $0.89$ & $32$ \\
DermaMNIST & global   & $+0.76$ & $0.41$ & $1.14$ & $53$ \\
DermaMNIST & tailored & $+0.44$ & $0.32$ & $0.88$ & $32$ \\
OCTMNIST   & oracle   & $+2.42$ & $0.17$ & $0.49$ & $10$ \\
OCTMNIST   & global   & $+0.13$ & $0.11$ & $0.31$ & $4$ \\
OCTMNIST   & tailored & $+0.15$ & $0.13$ & $0.37$ & $6$ \\
CIFAR-10   & oracle   & $+6.79$ & $0.21$ & $0.58$ & $14$ \\
CIFAR-10   & global   & $-0.05$ & $0.24$ & $0.66$ & $18$ \\
CIFAR-10   & tailored & $-0.23$ & $0.18$ & $0.50$ & $10$ \\
\bottomrule
\end{tabular}
\end{table}

\subsection{Volume scan and diagnostic dissociation}

Records behind Observation~\ref{prop:basisfirst} items (iii) and (v).

\begin{table}[!t]
\centering
\caption{Critical misaligned volume $K^{\star}$ from the dense scan
$K \in \{8, \dots, 512\}$ (first $K$ with relative damage above $5\%$),
against the input-layer tail exponent $\beta_o$ of the private bulk (the
scan is performed at the input layer). Damage beyond $K^{\star}$ grows
superlinearly on steep spectra and is bounded by the Haar-basis asymptote.}
\label{tab:volume}
\small
\setlength{\tabcolsep}{4pt}
\begin{tabular}{lccc}
\toprule
modality & $\beta_o$ (input) & $K^{\star}$ & res. at $K{=}512$ \\
\midrule
OCTMNIST     & $-2.23$ & $128$        & $32.6\times$ \\
DermaMNIST   & $-2.06$ & $128$        & $54.0\times$ \\
TissueMNIST  & $-1.78$ & $192$        & $6.0\times$ \\
CIFAR-10     & $-1.58$ & $320$        & $1.9\times$ \\
PathMNIST    & $-1.08$ & not reached  & $1.0\times$ \\
\bottomrule
\end{tabular}
\end{table}

\begin{figure}[!t]
\centering
\includegraphics[width=0.74\columnwidth]{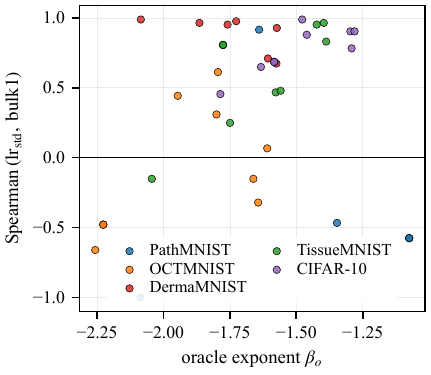}
\caption{Eigenvalue diagnostics invert in the steep regime: layer-wise
Spearman correlation between spectral dispersion and the bulk residual over
the extended probe grid. Negatives concentrate at flat inputs, overly steep
inputs, and steep output layers.}
\label{fig:dissoc}
\end{figure}

\subsection{BBP threshold sweep}

Injected-spike sweep behind Table~\ref{tab:bbp} and
Proposition~\ref{prop:t2} ($d = 784$, $n = 4096$, $d/n = 0.191$,
exact $s_c = 0.551$, leading order $0.661$): the ratio of the sample top
eigenvalue to the bulk edge, mean over $3$ seeds.

\begin{table}[!t]
\centering
\caption{Injected-spike sweep ($d = 784$, $n = 4096$, $d/n = 0.191$,
exact threshold $s_c = 0.551$, leading order $c^{1/4} = 0.661$): ratio of the
sample top eigenvalue to the bulk edge, mean over $3$ seeds. The empirical
crossing sits between $s = 0.6$ and $s = 0.7$, above the asymptotic threshold,
the upward shift expected at finite $n$ (Figure~\ref{fig:boundary}b); at $s \ge 1$ the population
covariance is rank-one and the ratio saturates at $d$.}
\label{tab:bbp}
\small
\setlength{\tabcolsep}{3pt}
\begin{tabular}{lccccc}
\toprule
$s$ & $0.0$ & $0.25$ & $0.5$ & $0.6$ & $0.7$ \\
top-1/bulk & $1.004$ & $1.004$ & $1.005$ & $1.012$ & $1.126$ \\
\midrule
$s$ & $0.75$ & $0.85$ & $1.0$ & $1.25$ & $1.5$ \\
top-1/bulk & $1.259$ & $1.865$ & $783.8$ & $783.8$ & $783.8$ \\
\bottomrule
\end{tabular}
\end{table}

The same statistic on real factors, to test whether the ruler transfers.
Protocol identical to the sweep above: one MLP per modality, three epochs on
$20$k samples, factors estimated on a fresh $4096$-sample draw, $n = 4096$
throughout, exact thresholds $s_c = 0.551$ at width $784$ and $0.577$ at width
$1024$ (the leading order $c^{1/4}$ overstates both). Two readings. First, on
the centered activation covariance the ratio of the two leading eigenvalues
runs $1.13$ to $25.5$ over fifteen modality--depth pairs and reads, at layers
$0/2/4$, PathMNIST $25.50/22.31/9.85$, TissueMNIST $4.92/1.93/3.48$,
DermaMNIST $2.58/1.16/1.47$, CIFAR-10 $2.41/2.09/1.96$ and OCTMNIST
$1.73/1.13/1.38$; the uncentered factor carries the mean-activation direction
as well and gives $1.15$ to $25.5$. Second, and this is why the first reading is
reported as a separation rather than as a spike strength, the same second
eigenvalue that reproduces the Marcenko--Pastur edge to $0.99$ on the injected
model sits $5.3$ to $73$ times above the edge implied by the bulk mean on the
real factors. A real activation spectrum is therefore not a Marcenko--Pastur
bulk with an isolated spike, and reading a real $\beta$ against $s_c$ would
need a bulk model these spectra satisfy.

\subsection{Dataset and architecture universality}

Input-layer bulk residual $bulk_1$ by probe family over eight
dataset configurations (mean over two probe seeds; oracles are 3-epoch MLPs
trained on $12$k samples). psdF: private Fourier PSD profile in the Fourier
basis; psdR: the same profile in a Haar-random basis; eigR: oracle eigenvalue
spectrum in a Haar basis; white: $\alpha=0$ colored noise.

\begin{table}[!t]
\centering
\caption{Universality across eight dataset configurations
(Section~\ref{sec:univ}). Each configuration uses a single trained network and
two independent probe bases, averaged here; the two bases differ from their
mean by at most $5.7\%$, which is the uncertainty of the psdR and eigR columns
(the psdF and white columns are stable below $1\%$). That spread does not
change the ordering of the columns nor the reported range.}
\label{tab:univ}
\small
\setlength{\tabcolsep}{3pt}
\begin{tabular}{lrrrrr}
\toprule
dataset & psdF & psdR & eigR & white & psdR/psdF \\
\midrule
PathMNIST      & $0.981$ & $2.22$  & $2.19$  & $0.981$ & $2.3\times$ \\
OCTMNIST       & $1.043$ & $56.55$ & $53.78$ & $1.399$ & $54.2\times$ \\
DermaMNIST     & $0.968$ & $86.89$ & $80.87$ & $1.097$ & $89.8\times$ \\
TissueMNIST    & $0.932$ & $30.36$ & $28.78$ & $1.025$ & $32.6\times$ \\
CIFAR-10 (32px) & $0.941$ & $31.11$ & $29.73$ & $1.012$ & $33.1\times$ \\
CIFAR-100 (10 cls) & $0.972$ & $18.67$ & $19.52$ & $0.982$ & $19.2\times$ \\
CIFAR-10 (50k)    & $0.957$ & $27.63$ & $27.17$ & $0.996$ & $28.9\times$ \\
DermaMNIST-64 & $0.966$ & $86.69$ & $80.96$ & $1.097$ & $89.8\times$ \\
\bottomrule
\end{tabular}
\end{table}

\ifLong\else
\begin{figure*}[!t]
\centering
\includegraphics[width=0.73\textwidth]{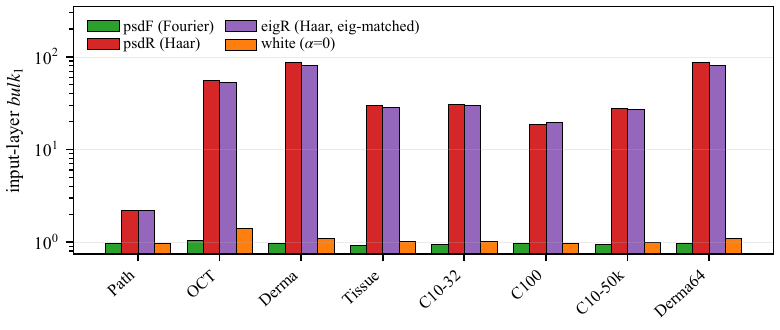}
\caption{Dataset sweep behind Table~\ref{tab:univ}: input-layer bulk
residual by probe family over eight dataset configurations. The
Fourier-aligned profile (psdF) tracks the white baseline on every
configuration; the Haar-basis variants separate on all but the near-isotropic
PathMNIST input spectrum.}
\label{fig:univdata}
\end{figure*}
\fi

Architecture control: input-layer bulk residual on three datasets
for an MLP oracle ($[784,256,256,10]$) and a conv oracle
(Conv$16$--Conv$32$--Linear; conv KFAC factors via unfold). The conv
input layer is conv1; the MLP input layer is layer 0.

\begin{table}[!t]
\centering
\caption{Architecture control (Section~\ref{sec:univ}). Each MLP row uses
the input layer, each CNN row the first convolutional layer, so the comparison
holds depth in the stack fixed rather than layer index.}
\label{tab:univarch}
\small
\setlength{\tabcolsep}{3pt}
\begin{tabular}{llrrrr}
\toprule
dataset & arch & psdF & psdR & white & psdR/psdF \\
\midrule
PathMNIST  & MLP & $0.981$ & $2.22$  & $0.981$ & $2.3\times$ \\
PathMNIST  & CNN & $0.120$ & $0.494$ & $0.497$ & $4.1\times$ \\
DermaMNIST & MLP & $0.968$ & $86.89$ & $1.097$ & $89.8\times$ \\
DermaMNIST & CNN & $0.119$ & $0.668$ & $0.671$ & $5.6\times$ \\
CIFAR-100  & MLP & $0.972$ & $18.67$ & $0.982$ & $19.2\times$ \\
CIFAR-100  & CNN & $0.099$ & $0.607$ & $0.610$ & $6.2\times$ \\
\bottomrule
\end{tabular}
\end{table}

\ifLong\else
\begin{figure}[!t]
\centering
\includegraphics[width=\columnwidth]{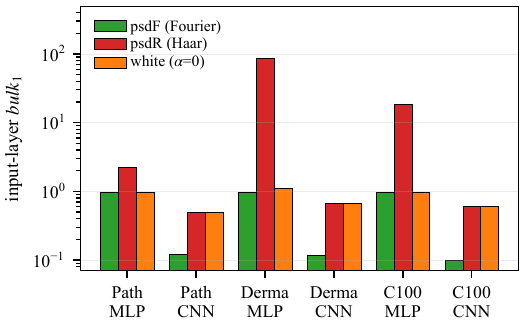}
\caption{Architecture sweep behind Table~\ref{tab:univarch}: on the
convolutional oracle the Haar-basis probe collapses onto the white-noise
baseline while the Fourier-aligned probe sits at one quarter to one sixth of
it; on the MLP oracle the Haar mismatch amplifies further.}
\label{fig:univarch}
\end{figure}
\fi

\subsection{Measurement pitfalls}
\label{app:pitfalls}

\textbf{Final-epoch accuracy is an artifact on imbalanced validation sets.}
On DermaMNIST ($66.9\%$ majority class, seven classes) per-epoch accuracy
oscillates between a chance-level basin and a below-chance basin, so any
single-epoch statistic is a coin flip. The diagnostic sweeps reported in this appendix use
balanced accuracy with a mean-of-last-three-epochs statistic over four seeds;
on this modality that statistic itself stays at chance for every probe
($0.100$--$0.118$ against the $0.10$ constant-predictor level of this
convention), so
no exponent ordering is measurable under it. The failure inverts conclusions:
under the fixed protocol the raw mean-of-last-three accuracy appears to order
with the bulk metric ($\alpha{=}2$: $0.644$, $\alpha{=}1$: $0.596$,
$\alpha{=}0$: $0.485$), the opposite of the ordering predicted by the
spike-inclusive total metric, yet that apparent ordering is majority-class
collapse rather than utility.

\textbf{A methodological inversion.} Within the steep regime the
spectral-dispersion diagnostic recommended by the shape hypothesis
anti-predicts the bulk residual on $12$ of $43$ layer-configurations,
including a perfect inversion (monotone residual increase with monotone
dispersion decrease). Diagnostics should be validated against the quantity
they claim to proxy; on this problem the eigenvalue layer is the wrong
abstraction.

\subsection{Monte-Carlo check of the whitened-volume law}

Proposition~\ref{prop:volume} was checked numerically on synthetic spectra
($d = 60$ and $d = 400$, $400$ Haar rotations per configuration). The
empirical mean of $\operatorname{tr}\M$ matches
$(\operatorname{tr}\Lambda)(\operatorname{tr}\Lambda^{-1})/d$ to within
Monte-Carlo error ($0.3\%$--$0.8\%$ across power laws
$\lambda_j \propto j^{-\gamma}$, $\gamma \in \{0.5, 1, 1.5, 2.5\}$);
$\det\M = 1$ holds to machine precision; the partial-block identity of
Lemma~\ref{lem:partial} matches to $0.2\%$--$1.3\%$; and at a fixed $5\%$-of-$d$ damage
threshold the critical volume decreases monotonically with steepness
($K^{\star} = 36, 16, 12, 8$ for $\gamma = 1.0, 1.5, 2.0, 2.2$), the same
ordering as Table~\ref{tab:volume}.

\ifLong\else
\subsection{Noise-tolerance sweep}

Mean-of-last-three-epochs test accuracy for the cross-over sweep of
Section~\ref{sec:negative} (PathMNIST, four seeds, learning rate $0.03$;
majority-class rate $0.143$). The matched sources lead below $\sigma = 0.3$;
the flat white probe leads above it.

\begin{table}[!t]
\centering
\caption{Noise tolerance and the source-ordering cross-over (four seeds).}
\label{tab:exp21}
\small
\setlength{\tabcolsep}{4pt}
\begin{tabular}{lcccc}
\toprule
$\sigma$ & $\varepsilon$ & private & aligned & white \\
\midrule
$0.0$ & --        & $0.322$ & $0.329$ & $0.130$ \\
$0.1$ & $1.1{\times}10^{4}$ & $0.217$ & $0.222$ & $0.132$ \\
$0.2$ & $274$     & $0.155$ & $0.157$ & $0.134$ \\
$0.3$ & $86$      & $0.126$ & $0.132$ & $\mathbf{0.155}$ \\
$0.5$ & $16.6$    & $0.110$ & $0.114$ & $\mathbf{0.128}$ \\
\bottomrule
\end{tabular}
\end{table}

\begin{figure}[!t]
\centering
\includegraphics[width=0.92\columnwidth]{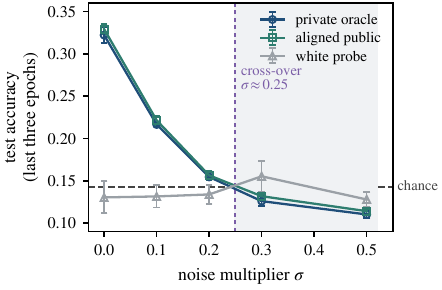}
\caption{Source-ordering cross-over of Table~\ref{tab:exp21} (PathMNIST, four
seeds, learning rate $0.03$, mean of the last three epochs; bars are
$\pm 1$ s.e.). Below $\sigma \approx 0.25$ the matched sources lead the
flat white probe; above it the order inverts, and at $\sigma = 0.3$ the
white probe is the only arm above the majority-class rate (dashed).}
\label{fig:crossover}
\end{figure}

\subsection{The $\varepsilon \le 10$ workspace}

Ten seeds per arm, $\sigma = 0.5$, batch $256$, twelve epochs on the full
training split of $89996$ samples, $4.5\times$ the $20000$ used in
Section~\ref{sec:negative}. Widening the set is the only lever that reaches a
budget below ten at fixed noise: with $\delta = 1/N$ the RDP accountant gives
$\varepsilon = 9.99$. The three real-source arms read $0.125$, $0.130$ and
$0.125$ (mean of the last three epochs, balanced), the two synthetic arms
$0.144$ and $0.140$, and the constant predictor $0.143$: no real-source arm
clears it and the two synthetic arms sit on it. The cross-over therefore does
not move into $\varepsilon \le 10$ under this widening, and the binding
constraint is the noise rather than the sample size.
\fi

\ifLong
\clearpage
\bibliography{refs}

\begin{thebibliography}{28}
\providecommand{\natexlab}[1]{#1}
\providecommand{\url}[1]{#1}
\csname url@samestyle\endcsname
\providecommand{\newblock}{\relax}
\providecommand{\bibinfo}[2]{#2}
\providecommand{\BIBentrySTDinterwordspacing}{\spaceskip=0pt\relax}
\providecommand{\BIBentryALTinterwordstretchfactor}{4}
\providecommand{\BIBentryALTinterwordspacing}{\spaceskip=\fontdimen2\font plus
\BIBentryALTinterwordstretchfactor\fontdimen3\font minus
  \fontdimen4\font\relax}
\providecommand{\BIBforeignlanguage}[2]{{%
\expandafter\ifx\csname l@#1\endcsname\relax
\typeout{** WARNING: IEEEtranN.bst: No hyphenation pattern has been}%
\typeout{** loaded for the language `#1'. Using the pattern for}%
\typeout{** the default language instead.}%
\else
\language=\csname l@#1\endcsname
\fi
#2}}
\providecommand{\BIBdecl}{\relax}
\BIBdecl

\bibitem[Dwork and Roth(2014)]{dwork2014}
C.~Dwork and A.~Roth, ``The algorithmic foundations of differential privacy,''
  \emph{Foundations and Trends in Theoretical Computer Science}, vol.~9, no.
  3--4, pp. 211--407, 2014.

\bibitem[Abadi et~al.(2016)Abadi, Chu, Goodfellow, McMahan, Mironov, Talwar,
  and Zhang]{abadi2016}
M.~Abadi, A.~Chu, I.~Goodfellow, H.~B. McMahan, I.~Mironov, K.~Talwar, and
  L.~Zhang, ``Deep learning with differential privacy,'' in \emph{Proceedings
  of the 2016 ACM SIGSAC Conference on Computer and Communications Security},
  2016, pp. 308--318.

\bibitem[Dauphin et~al.(2014)Dauphin, Pascanu, Gulcehre, Cho, Ganguli, and
  Bengio]{dauphin2014}
Y.~N. Dauphin, R.~Pascanu, C.~Gulcehre, K.~Cho, S.~Ganguli, and Y.~Bengio,
  ``Identifying and attacking the saddle point problem in high-dimensional
  non-convex optimization,'' \emph{Advances in Neural Information Processing
  Systems}, vol.~27, 2014.

\bibitem[Ganesh et~al.(2025)Ganesh, McMahan, and Thakurta]{ganesh2025}
A.~Ganesh, B.~McMahan, and A.~Thakurta, ``On design principles for private
  adaptive optimizers,'' \emph{arXiv preprint arXiv:2507.01129, PPML 2025},
  2025.

\bibitem[Krouka et~al.(2025)Krouka, Koskela, and Kulkarni]{krouka2025}
M.~Krouka, A.~Koskela, and T.~Kulkarni, ``Communication efficient
  differentially private federated learning using second order information,''
  \emph{Proceedings on Privacy Enhancing Technologies}, vol. 2025, no.~1, pp.
  584--612, 2025.

\bibitem[Amid et~al.(2022)Amid, Ganesh, Mathews, Ramaswamy, Song, Steinke,
  Suriyakumar, Thakkar, and Thakurta]{amid2022}
E.~Amid, A.~Ganesh, R.~Mathews, S.~Ramaswamy, S.~Song, T.~Steinke, V.~M.
  Suriyakumar, O.~Thakkar, and A.~Thakurta, ``Public data-assisted mirror
  descent for private model training,'' in \emph{Proceedings of the 39th
  International Conference on Machine Learning}, ser. Proceedings of Machine
  Learning Research, vol. 162.\hskip 1em plus 0.5em minus 0.4em\relax PMLR,
  2022, pp. 517--535.

\bibitem[Molina Van~den Bosch et~al.(2026)Molina Van~den Bosch, Taiello,
  Sund~Aillet, Protani, Gonzalez~Ballester, and Serio]{dpkfc2026}
M.~Molina Van~den Bosch, R.~Taiello, A.~Sund~Aillet, A.~Protani, M.~A.
  Gonzalez~Ballester, and L.~Serio, ``{DP-KFC}: Data-free preconditioning for
  privacy-preserving deep learning,'' \emph{arXiv preprint arXiv:2605.13418},
  2026.

\bibitem[Martens and Grosse(2015)]{kfac2015}
J.~Martens and R.~Grosse, ``Optimizing neural networks with kronecker-factored
  approximate curvature,'' in \emph{Proceedings of the 32nd International
  Conference on Machine Learning}, ser. Proceedings of Machine Learning
  Research, vol.~37.\hskip 1em plus 0.5em minus 0.4em\relax PMLR, 2015, pp.
  2408--2417.

\bibitem[Mar{\v{c}}enko and Pastur(1967)]{marchenko1967}
V.~A. Mar{\v{c}}enko and L.~B. Pastur, ``Distribution of eigenvalues for some
  sets of random matrices,'' \emph{Matematicheskii Sbornik}, vol.~72, no. 114,
  pp. 507--536, 1967, english translation: Mathematics of the USSR-Sbornik,
  1(4):457--483.

\bibitem[Johnstone(2001)]{johnstone2001}
I.~M. Johnstone, ``On the distribution of the largest eigenvalue in principal
  components analysis,'' \emph{The Annals of Statistics}, vol.~29, no.~2, pp.
  295--327, 2001.

\bibitem[Baik et~al.(2005)Baik, Ben~Arous, and P{\'e}ch{\'e}]{baik2005}
J.~Baik, G.~Ben~Arous, and S.~P{\'e}ch{\'e}, ``Phase transition of the largest
  eigenvalue for nonnull complex sample covariance matrices,'' \emph{The Annals
  of Probability}, vol.~33, no.~5, pp. 1643--1697, 2005.

\bibitem[Wang et~al.(2024)Wang, Wu, and Fan]{wang2024}
Z.~Wang, D.~Wu, and Z.~Fan, ``Nonlinear spiked covariance matrices and signal
  propagation in deep neural networks,'' in \emph{Proceedings of the 41st
  International Conference on Machine Learning}, ser. Proceedings of Machine
  Learning Research, vol. 235, 2024, arXiv:2402.10127.

\bibitem[Sagun et~al.(2016)Sagun, Bottou, and LeCun]{sagun2016}
L.~Sagun, L.~Bottou, and Y.~LeCun, ``Eigenvalues of the hessian in deep
  learning: Singularity and beyond,'' \emph{arXiv preprint arXiv:1611.07476},
  2016.

\bibitem[Yang et~al.(2023)Yang, Shi, Wei, Liu, Zhao, Ke, Pfister, and
  Ni]{medmnist2023}
J.~Yang, R.~Shi, D.~Wei, Z.~Liu, L.~Zhao, B.~Ke, H.~Pfister, and B.~Ni,
  ``Medmnist v2: A large-scale lightweight benchmark for 2d and 3d biomedical
  image classification,'' \emph{Scientific Data}, vol.~10, no.~1, p.~41, 2023.

\bibitem[Yang et~al.(2021)Yang, Shi, and Ni]{medmnist2021}
J.~Yang, R.~Shi, and B.~Ni, ``Medmnist classification decathlon: A lightweight
  automl benchmark for medical image analysis,'' in \emph{IEEE 18th
  International Symposium on Biomedical Imaging (ISBI)}, 2021, pp. 191--195.

\bibitem[Krizhevsky(2009)]{krizhevsky2009}
A.~Krizhevsky, ``Learning multiple layers of features from tiny images,''
  University of Toronto, Tech. Rep., 2009.

\bibitem[Couillet and Liao(2022)]{couillet2022}
R.~Couillet and Z.~Liao, \emph{Random Matrix Methods for Machine
  Learning}.\hskip 1em plus 0.5em minus 0.4em\relax Cambridge University Press,
  2022.

\bibitem[Liao and Mahoney(2026)]{liao2026}
Z.~Liao and M.~W. Mahoney, ``Random matrix theory for deep learning: Beyond
  eigenvalues of linear models,'' \emph{IEEE Signal Processing Magazine},
  vol.~43, no.~2, pp. 93--106, 2026, arXiv:2506.13139.

\bibitem[Jung et~al.(2023)Jung, Chung, and Lee]{jung2023}
J.~H. Jung, H.~W. Chung, and J.~O. Lee, ``Detection problems in the spiked
  matrix models,'' \emph{arXiv preprint arXiv:2301.05331}, 2023.

\bibitem[Fan and Ma(2026)]{fan2026}
Z.~Fan and R.~Ma, ``Kronecker-product random matrices and a matrix least
  squares problem,'' \emph{The Annals of Probability}, vol.~54, no.~2, pp.
  973--1033, 2026, arXiv:2406.00961.

\bibitem[Chen et~al.(2026)Chen, Zhang, Liu, Li, Sun, Li, Zhang, Han, Tan, and
  Chen]{zeta2026}
K.~Chen, S.~Zhang, Z.~Liu, L.~Li, Y.~Sun, Y.~Li, Y.~Zhang, B.~Han, M.~Tan, and
  Q.~Chen, ``Zeta: Dual whitening for matrix optimization via
  coordinate-adaptive preconditioning,'' \emph{arXiv preprint
  arXiv:2606.14187}, 2026.

\bibitem[Seung et~al.(2025)Seung, Lee, and Ko]{mac2025}
H.~Seung, J.~Lee, and H.~Ko, ``{MAC}: An efficient gradient preconditioning
  using mean activation approximated curvature,'' in \emph{IEEE International
  Conference on Data Mining (ICDM)}, 2025, arXiv:2506.08464.

\bibitem[Xia et~al.(2025)Xia, Zhang, and Cai]{xia2025}
X.~Xia, L.~Zhang, and Z.~Cai, ``Statistical inference for differentially
  private stochastic gradient descent,'' \emph{arXiv preprint
  arXiv:2507.20560}, 2025.

\bibitem[Poole et~al.(2016)Poole, Lahiri, Raghu, Sohl-Dickstein, and
  Ganguli]{poole2016}
B.~Poole, S.~Lahiri, M.~Raghu, J.~Sohl-Dickstein, and S.~Ganguli, ``Exponential
  expressivity in deep neural networks through transient chaos,'' in
  \emph{Advances in Neural Information Processing Systems}, vol.~29, 2016.

\bibitem[Schoenholz et~al.(2017)Schoenholz, Gilmer, Ganguli, and
  Sohl-Dickstein]{schoenholz2017}
S.~S. Schoenholz, J.~Gilmer, S.~Ganguli, and J.~Sohl-Dickstein, ``Deep
  information propagation,'' in \emph{International Conference on Learning
  Representations}, 2017.

\bibitem[Benaych-Georges and Nadakuditi(2011)]{bg2011}
F.~Benaych-Georges and R.~R. Nadakuditi, ``The eigenvalues and eigenvectors of
  finite, low rank perturbations of large random matrices,'' \emph{Advances in
  Mathematics}, vol. 227, no.~1, pp. 494--521, 2011, arXiv:0910.2120.

\bibitem[Bak et~al.(1987)Bak, Tang, and Wiesenfeld]{bak1987}
P.~Bak, C.~Tang, and K.~Wiesenfeld, ``Self-organized criticality: An
  explanation of $1/f$ noise,'' \emph{Physical Review Letters}, vol.~59, no.~4,
  pp. 381--384, 1987.

\bibitem[Friston(2010)]{friston2010}
K.~Friston, ``The free-energy principle: a unified brain theory?'' \emph{Nature
  Reviews Neuroscience}, vol.~11, no.~2, pp. 127--138, 2010.

\end{thebibliography}
\bibliographystyle{IEEEtranN}
\fi

\end{document}